\documentclass{article}

\usepackage{arxiv}
\usepackage[authoryear,longnamesfirst]{natbib}
\usepackage[utf8]{inputenc} 
\usepackage[T1]{fontenc}   
\usepackage{hyperref}       
\usepackage{url}            
\usepackage{booktabs}       
\usepackage{amsfonts}       
\usepackage{nicefrac}       
\usepackage{microtype}      
\usepackage{lipsum}
\usepackage{graphicx}
\usepackage{acronym}
\usepackage{textcomp}
\usepackage{xurl}
\usepackage{amsmath}
\usepackage{caption}
\usepackage{subcaption}
\usepackage{multirow}

\graphicspath{ {./images/} }

\acrodef{UCE}{Uncertainty Calibration Error}
\acrodef{ECE}{Expected Calibration Error}
\acrodef{BCCE}{Boundary Curve Calibration Error}
\acrodef{DTS}{Dual Temperature Scaling}
\acrodef{TS}{Temperature Scaling}
\acrodef{CUBC}{Confidence-Uncertainty Boundary Curve}
\acrodef{BNN}{Bayesian Neural Network}
\acrodef{MC}{Monte Carlo}
\acrodef{VI}{Variational Inference}
\acrodef{ELBO}{Evidence Lower Bound}
\acrodef{CUB-Loss}{Confidence-Uncertainty Boundary Loss}
\acrodef{ID}{In-Distribution}
\acrodef{OOD}{Out-of-Distribution}
\acrodef{AvU}{Accuracy vs. Uncertainty}
\acrodef{AvUC}{Accuracy vs. Uncertainty Calibration}
\acrodef{AUPR}{Area Under the Precision-Recall Curve}
\acrodef{AUROC}{Area Under the Receiver Operating Characteristic Curve}
\acrodef{BACC}{Balanced Accuracy}
\acrodef{SGD}{Stochastic Gradient Descent}
\acrodef{MOPED}{Model Priors with Empirical Bayes using Deterministic Neural Networks}
\acrodef{HAM10000}{Human Against Machine with 10000 Training Images}
\acrodef{DR}{Diabetic Retinopathy}
\acrodef{NV}{Melanocytic nevi}
\acrodef{MEL}{Melanoma}
\acrodef{BKL}{Benign keratosis-like lesions}
\acrodef{BCC}{Basal cell carcinoma}
\acrodef{AKIEC}{Actinic keratoses}
\acrodef{VASC}{Vascular lesions}
\acrodef{DF}{Dermatofibroma}
\acrodef{AC}{Accurate-Certain}
\acrodef{IU}{Inaccurate-Uncertain}
\acrodef{IC}{Inaccurate-Certain}
\acrodef{AU}{Accurate-Uncertain}
\acrodef{PCA}{Principal Component Analysis}
\acrodef{RFF}{Random Fourier Features}

\title{Confidence--Uncertainty Boundary Calibration for Bayesian Deep Learning in Medical Image Analysis}

\author{
  Hua Xu \\
  Escuela Técnica Superior de Ingenieros de Telecomunicación\\
  Universidad Politécnica de Madrid\\
  28040, Madrid, Spain\\ 
  \texttt{hua.xu@alumnos.upm.es} \\
   \And
 Julián D. Arias-Londoño \\
  Escuela Técnica Superior de Ingenieros de Telecomunicación\\
  Universidad Politécnica de Madrid\\
  28040, Madrid, Spain\\ 
  \texttt{julian.arias@upm.es} \\
  \And
 Juan I. Godino-Llorente \\
  Escuela Técnica Superior de Ingenieros de Telecomunicación\\
  Universidad Politécnica de Madrid\\
  28040, Madrid, Spain\\ 
  \texttt{ignacio.godino@upm.es} \\
}

\begin{document}
\maketitle
\begin{abstract}
In critical decision support systems based on medical imaging, the reliability of AI-assisted decision-making is as relevant as predictive accuracy. Although deep learning models have demonstrated significant accuracy, they frequently suffer from miscalibration, manifested as overconfidence in erroneous predictions. To facilitate clinical acceptance, it is imperative that models quantify uncertainty in a manner that correlates with prediction correctness, allowing clinicians to identify unreliable outputs for further review. In order to address this necessity, the present paper proposes a probabilistic optimization framework grounded in Bayesian deep learning. Specifically, the \ac{CUBC} is first explored as an intermediate operational target. Grounded in this target, a novel \ac{CUB-Loss} is proposed to regularize the alignment between prediction confidence and uncertainty estimates during training, imposing penalties on high-certainty errors and low-certainty correct predictions. Upon completion of training optimization, a \ac{BCCE} metric is further introduced to measure the degree of boundary alignment in the calibrated model. Building on this measurement, a \ac{DTS} strategy is devised to perform post-hoc refinement, further adjusting the posterior predictive distribution across different confidence--uncertainty regions. The proposed framework is validated on three distinct medical imaging tasks: automatic screening of pneumonia, diabetic retinopathy detection, and identification of skin lesions. Empirical results demonstrate that the proposed approach improves uncertainty calibration across diverse modalities, maintains robust performance in data-scarce scenarios, and remains effective on severely imbalanced datasets, underscoring its potential for real clinical deployment.
\end{abstract}

\keywords{Uncertainty Calibration \and Confidence--Uncertainty Boundary \and Bayesian Deep Learning \and Dual Temperature Scaling \and Medical Image Analysis}

\acresetall

\section{Introduction}
\label{sec:introduction}

Deep learning has achieved remarkable advances in medical image analysis, demonstrating classification performance comparable to that of human experts across many application domains. Predominantly, these approaches rely on deterministic deep neural networks, which operate on the maximum likelihood estimation principle by optimizing weights to minimize the negative log-likelihood loss. However, while effective in achieving high accuracy, this paradigm is prone to overfitting, often resulting in overconfidence~\citep{guo2017calibration}. This overconfidence obscures the distinction between reliable and unreliable predictions, leaving clinicians without a principled basis for deciding when to trust model outputs. Consequently, in critical healthcare applications, models must not only achieve high predictive accuracy but also quantify their predictive uncertainty, providing clinicians with an explicit indicator of output reliability to support informed decision-making~\citep{leibig2017leveraging,combalia2020uncertainty}.

In this context, \acp{BNN} offer a framework for uncertainty quantification by modeling weight distributions rather than point estimates~\citep{mackay1992bayesian,neal1996bayesian}. Two practical approximation families are especially relevant here: \ac{MC} Dropout~\citep{gal2016dropout} and \ac{VI}~\citep{blundell2015weight, graves2011practical}. \ac{MC} Dropout offers computational convenience and remains a useful stochastic Bayesian baseline, but its approximate posterior is induced by dropout design choices such as mask placement and dropout rate~\citep{gal2017concrete}. In contrast, \ac{VI} explicitly approximates the posterior distribution of weights by optimizing the \ac{ELBO}, allowing the variance of each weight parameter to be learned from data~\citep{blundell2015weight}.  This approach facilitates a more expressive and fine-grained quantification of uncertainty, providing the rigorous probabilistic foundation required for high-stakes medical decision-making~\citep{combalia2020uncertainty}.

However, \acsp{BNN} exhibit practical limitations in real-world medical imaging applications~\citep{kurz2022uncertainty}. In particular, their predictive performance and uncertainty estimates often degrade under small-sample and severely imbalanced data regimes, which are prevalent in clinical datasets~\citep{mallick2020can}. Moreover, although \acsp{BNN} offer a principled framework for uncertainty quantification, the resulting uncertainty estimates are not inherently guaranteed to be decision-consistent, i.e., systematically low for correct predictions and high for incorrect ones, unless explicitly enforced through additional modeling or optimization constraints~\citep{krishnan2020improving}.

These limitations highlight the need for a unified optimization framework that explicitly enforces alignment between predictive correctness and uncertainty, while maintaining robustness across varying data scales, class distributions, and clinical tasks. Such a framework is essential to ensure both the reliability and general applicability of uncertainty-aware models in diverse medical imaging scenarios.

To address these challenges, this work proposes a generalizable probabilistic optimization framework grounded in \acp{BNN}. Rather than relying on task-specific architectural modifications, the focus is placed on the fundamental training objective to ensure broad applicability. Specifically, a novel \ac{CUBC} is introduced, which establishes an explicit functional mapping between predictive confidence and uncertainty. Building upon this theoretical foundation, the \ac{CUB-Loss} is proposed. This loss operates as a semantic bridge, explicitly aligning the model's uncertainty estimates with human cognitive patterns, where high confidence corresponds to low uncertainty and high error risk corresponds to high uncertainty. By imposing these directional constraints during optimization, the proposed \ac{CUB-Loss} anchors the magnitude of uncertainty to prediction correctness, effectively narrowing the gap between raw model outputs and human-intelligible uncertainty signals. Complementing this training-time optimization, a \ac{DTS} strategy is proposed to fine-tune the posterior distribution, further enhancing the calibration alignment to provide clinicians with an explicit reference for determining the acceptance of model predictions.

The main contributions of this paper are summarized as follows:

\begin{itemize}
    \item The formulation of a Bayesian framework that utilizes \ac{VI} with \ac{MC} sampling to capture uncertainty. Central to this framework is the proposal of the \ac{CUBC}, which establishes a theoretical mapping between confidence and uncertainty. Based on this foundation, \ac{CUB-Loss} is proposed as a novel objective function that strengthens the boundary-aware training signal derived from the \ac{CUBC}, enforcing the coupling between prediction correctness and uncertainty estimates to enhance model calibration effectively.

    \item The introduction of the \ac{BCCE}, a novel metric that quantifies the discrepancy between the model's uncertainty estimates and the theoretical \ac{CUBC}, providing an explicit diagnostic tool for evaluating post-training boundary alignment. Informed by this measurement, a post-hoc calibration strategy termed \ac{DTS} is further devised, which employs bidirectional \ac{TS} with two distinct temperature parameters to perform region-specific temperature adjustment across different confidence--uncertainty regions. 

    \item A comprehensive evaluation on three representative medical imaging tasks: 1) the screening of different pneumonia types from chest radiographs; 2)
    the detection of \ac{DR} from fundus photographs;
    and, 3) the identification of skin lesions.
    They cover different scenarios about Near-\ac{OOD} detection, data scarcity learning, and severe class imbalance, and were selected to demonstrate the generalizability and robustness of the proposed framework. 

\end{itemize}

The paper is organized as follows: Section \ref{sec:related_work} presents the techniques typically used in the state of the art for posterior inference in \acp{BNN} and the metrics commonly used, which will be utilized throughout this work; Section \ref{sec:methodology} describes the methods used and those proposed in this paper; Section \ref{sec:experiments} presents a detailed account of the experiments carried out and the results obtained in three different application domains; Section~\ref{sec:discussion} synthesizes the findings across the three experiments, clarifies the complementary roles of the proposed components, and discusses future work; and finally, Section \ref{sec:conclusions} draws the main conclusions of the work.

\section{Related work}\label{sec:related_work}

This section presents an overview and discusses the limitations of the most relevant approximations found in the literature that are considered specifically relevant to the main objectives of this paper.  

\subsection{Posterior inference in Bayesian Neural Networks}
\label{sec:related_bnn}

Standard deterministic neural networks rely on point estimates, assigning a single fixed value to each weight parameter, which often leads to overconfident predictions. In contrast, \acp{BNN} adopts a probabilistic framework by treating model weights not as deterministic values, but as random variables governed by probability distributions. This formulation allows the model to output a predictive distribution rather than a single scalar, inherently capturing model uncertainty. 

However, exact posterior inference is analytically intractable for modern, high-dimensional neural architectures. To overcome this computational bottleneck, two primary approximation strategies have emerged: \ac{MC} Dropout and \ac{VI}.

\subsubsection{\acs{MC} Dropout}
\label{sec:mc_dropout}

\ac{MC} Dropout~\citep{gal2016dropout} interprets dropout training as a lightweight variational Bayesian approximation and has become a practical route for estimating predictive uncertainty without redesigning the full network. This makes it an important stochastic uncertainty baseline rather than a non-Bayesian alternative. However, the posterior approximated is implicitly defined by a fixed Bernoulli dropout mask, which imposes a rigid uncertainty structure across the network and limits its capacity to capture complex posterior landscapes. Empirical studies have confirmed this limitation. In this regard, \citep{djupskaas2026unreliable} demonstrate that \ac{MC} Dropout tends to produce uniform uncertainty across the input space, failing to capture increased uncertainty in extrapolation and interpolation regions.

Several approaches have attempted to address \ac{MC} Dropout's structural constraints at the algorithmic level. Variational Dropout~\citep{kingma2015variational} reinterprets Gaussian Dropout within a variational framework, enabling learnable dropout rates through the local reparameterization trick. Concrete Dropout~\citep{gal2017concrete} further extends this idea by employing continuous relaxations of discrete dropout masks, allowing automatic tuning of dropout probabilities during training. However, these methods remain fundamentally constrained by the dropout paradigm. Although Concrete Dropout enables automatic tuning of dropout probabilities, it parameterizes uncertainty through per-layer dropout rates rather than per-weight distributions ---the posterior still operates via stochastic masking rather than continuous density estimation over weight space.

Complementary to these structural improvements, numerous post-hoc calibration methods have been proposed to recalibrate \ac{MC} Dropout's uncertainty outputs~\citep{laves2019temperature,zeevi2025ratein,ledda2023dropout}. However, such approaches primarily operate at inference time and cannot overcome the fundamental posterior expressiveness constraints inherent to dropout-based approximations.

\subsubsection{Variational Inference}

In contrast to the aforementioned methods based on MC, \ac{VI}-based \acp{BNN}~\citep{blundell2015weight,graves2011practical} naturally exhibit increasing uncertainty in extrapolation regions~\citep{djupskaas2026unreliable}, reflecting the desirable Bayesian property of appropriately decreasing certainty in data-sparse areas. 
This desirable behavior is enabled by \ac{VI}'s capacity to place continuous distributions over weights with independently learnable mean and variance for each parameter, offering greater statistical flexibility for capturing parameter-specific uncertainty.

\subsubsection{Calibration deficiencies in Variational Inference}

Despite their theoretical advantages, recent research has uncovered calibration deficiencies in \ac{VI}-based \acp{BNN}, which directly motivates the present work. 
First, \ac{VI}-based \acp{BNN} exhibit degraded uncertainty calibration under distributional shift. In particular, large-scale evaluations demonstrate that \ac{VI}-based \acp{BNN} often become overconfident on \ac{OOD} inputs~\citep{ovadia2019can}, a behavior attributed to approximation-induced posterior contraction rather than Bayesian inference itself~\citep{foong2019between,farquhar2020radial}.

Moreover, the cold posterior phenomenon reveals that applying post-hoc or implicit \ac{TS} can substantially improve predictive performance and uncertainty behavior in \ac{VI}-trained \acp{BNN}, indicating that the standard \ac{ELBO} objective alone does not yield well-calibrated predictive distributions~\citep{wenzel2020good}. 

These findings suggest that \ac{VI}-based \acp{BNN} require explicit and reliable calibration mechanisms.

\subsection{Accuracy vs. uncertainty}
\label{sec:related_avu}

Recent advances in calibration have shifted the focus from minimizing distribution-level discrepancies to explicitly optimizing the joint behavior of prediction accuracy and uncertainty. Under this paradigm, a reliable model is expected to be certain when its predictions are correct and uncertain when its predictions are incorrect. This alignment ensures that the model's confidence scores serve as a reliable indicator of correctness, thereby establishing trustworthiness in safety-critical applications.

This intuition is formalized in~\citep{krishnan2020improving} through the \ac{AvU} framework. In this regard, the authors proposed a taxonomy that categorizes prediction outcomes into four mutually exclusive states based on the alignment between correctness and certainty. The \ac{AC} state represents ideal reliability where the model is both correct and certain. The \ac{IU} state represents valid uncertainty estimation where the model erroneously predicts but appropriately signals a lack of certainty. These two states are desirable and should be maximized. Conversely, the \ac{IC} state indicates high-risk behavior where the model commits errors with a high degree of certainty. The \ac{AU} state reflects excessive caution where the model is correct but fails to exhibit sufficient certainty. These latter two states constitute misalignment and should be minimized during training. Table~\ref{tab:avu_states} provides a summary of these four states.

\begin{table}[htbp]
    \centering
    \caption{Summary of the four prediction states in the AvU framework.}
    \label{tab:avu_states}
    \setlength{\tabcolsep}{10pt}
    \renewcommand{\arraystretch}{1.2}
    \begin{tabular}{l l}
    \toprule
    State & Components \\
    \midrule
    AC & Accurate + Certain  \\
    IU & Inaccurate + Uncertain  \\
    IC & Inaccurate + Certain  \\
    AU & Accurate + Uncertain  \\
    \bottomrule
    \end{tabular}
\end{table}

Assuming that $n_{\mathrm{AC}}$, $n_{\mathrm{AU}}$, $n_{\mathrm{IC}}$, and $n_{\mathrm{IU}}$ denote the number of samples in each respective state. The \ac{AvU} metric is defined as:
\begin{equation}
    \mathrm{AvU} = \frac{n_{\mathrm{AC}} + n_{\mathrm{IU}}}{n_{\mathrm{AC}} + n_{\mathrm{AU}} + n_{\mathrm{IC}} + n_{\mathrm{IU}}}
\end{equation}
This metric directly measures the proportion of samples that exhibit dependable behavior, providing an intuitive and intelligible measure of uncertainty-accuracy alignment.

To enable end-to-end optimization~\citep{krishnan2020improving} introduced the \ac{AvUC}, a differentiable surrogate defined as:
\begin{equation}
    \mathcal{L}_{\mathrm{AvUC}} := -\log \left( \frac{n_{\mathrm{AC}} + n_{\mathrm{IU}}}{n_{\mathrm{AC}} + n_{\mathrm{IU}} + n_{\mathrm{AU}} + n_{\mathrm{IC}}} \right)
\end{equation}
To make this objective differentiable, the discrete state counts are replaced with soft approximations using $\tanh(\cdot)$ functions. Specifically, for each sample $i$ with an uncertainty estimate $u_i$ and an uncertainty threshold $u_{\mathrm{th}}$, the soft counts are computed by weighting contributions with $\tanh(u_i)$ as a smooth indicator of uncertainty and $p_i$ as a soft indicator of prediction correctness. The subsequent work in~\citep{karandikar2021soft} proposed soft calibration objectives S-\ac{AvUC}, which further refines this relaxation through temperature-controlled soft binning to improve optimization stability.

Despite the conceptual appeal of the \ac{AvU} framework, existing optimization methods suffer from two fundamental limitations. First, the reliance on soft approximations for discrete state counts introduces a discrepancy between the surrogate objective and the actual \ac{AvU} metric. The $\tanh$-based relaxation merely smooths the counting operation but does not alter its categorical nature.

Second, the \ac{AvU} framework is inherently \emph{state-based} rather than \emph{distance-based}. It classifies samples into discrete categories but fails to quantify \emph{how far} a sample deviates from a dependable configuration. Consequently, a model might achieve a high \ac{AvU} score while still exhibiting suboptimal uncertainty magnitudes. For instance, two \ac{IC} samples contribute equally to the loss even if one has confidence marginally above the threshold and the other is severely overconfident. This categorical treatment obscures the severity of miscalibration and provides uniform gradients regardless of deviation magnitude.

Motivated by these observations, the approach proposed in this paper departs from count-based optimization of categorical states. Instead of relying on soft approximations to discrete counts, a \ac{CUBC} is derived that defines the theoretical relationship between confidence and uncertainty. This geometric formulation allows a distance-based loss to be defined that directly measures how far each sample deviates from the ideal boundary, providing gradients proportional to the severity of miscalibration.

\subsection{Post-hoc calibration and the limits of linear scaling}
\label{sec:related_ts}

Originally proposed in~\citep{guo2017calibration}, \ac{TS} remains the most prevalent post-hoc calibration method due to its simplicity and effectiveness. Ideally, a calibrated model should minimize the \ac{ECE}, which measures the discrepancy between predicted confidence and empirical accuracy. \ac{TS} addresses this by introducing a single scalar parameter $T>0$ to rescale the logits before the $softmax$ function. By optimizing $T$ to minimize the negative log-likelihood on a validation set, \ac{TS} effectively aligns the confidence distribution without affecting the model's classification accuracy.

Following this success, recent research has extended the concept of calibration from confidence to predictive uncertainty. In this regard,~\citep{laves2019temperature} introduced the \ac{UCE} to quantify the misalignment between estimated uncertainty and regression error, proposing that \ac{TS} could also be used to calibrate uncertainty estimates. However, a critical limitation of these approaches is the implicit assumption of a simplified or linear relationship between the scaling factor and the resulting uncertainty distribution. Standard \ac{TS} applies a global scaling operation that treats all samples uniformly, often neglecting the intrinsic, non-linear geometric relationship between confidence and uncertainty. The mere minimization of a scalar error metric like \ac{UCE} does not guarantee that the uncertainty values respect the theoretical bounds of the probability simplex.

This work addresses this gap by explicitly modeling the calibration problem relative to the theoretical \ac{CUBC}, rather than relying on the linear scaling assumption inherent in traditional \ac{UCE} optimization.

\section{Methodology}\label{sec:methodology}

This section presents the key concepts of the methods used and those proposed in this paper, starting from a brief introduction to \acp{BNN} and the notation followed, followed by the theoretical definition of the \ac{CUBC}, and ending with the proposed \ac{CUB-Loss} and \ac{DTS} strategy for post-hoc calibration.

\subsection{Preliminaries: Bayesian Neural Network}
\label{sec:preliminaries}

Standard computer-aided decision support systems typically employ deterministic deep neural networks. These models optimize a point estimate of the weights $w$ via the maximum likelihood estimation principle. However, this approach is fundamentally prone to overfitting and, crucially, lacks the capability to capture the uncertainty associated with the model's decisions ~\citep{mackay1992practical, mackay1992bayesian}.

To quantify uncertainty and provide probabilistic confidence level, the objective of \acp{BNN} is to compute the predictive probability distribution $p(y|x, \mathcal{D})$ ~\citep{neal1996bayesian, blundell2015weight}. This computation requires the marginalization over the posterior distribution of the weights $p(w|\mathcal{D})$:

\begin{equation}
    p(y|x, \mathcal{D}) = \int p(y|x, w) p(w|\mathcal{D}) \, dw
    \label{eq:predictive_integral}
\end{equation}
where $p(y|x, w)$ denotes the likelihood of the prediction given specific weights, and $dw$ indicates integration over the continuous weight space.

Given the evidence data $\mathcal{D} = \{(x, y)\}$, the prior distribution $p(w)$, and the model likelihood $p(y|x, w)$, the goal is to infer the posterior distribution over the weights $p(w|\mathcal{D})$. According to Bayes' theorem, this is formulated as:

\begin{equation}
    p(w|\mathcal{D}) = \frac{p(y|x, w)p(w)}{\int p(y|x, w)p(w) \, dw}
    \label{eq:posterior_calc}
\end{equation}

\begin{sloppy}
In this equation, the numerator contains the likelihood $p(y|x, w)$ and the prior $p(w)$, while the denominator $\int p(y|x, w)p(w) \, dw$ represents the marginal likelihood (model evidence), which serves as a normalizing constant.
\end{sloppy}

Computing the exact posterior $p(w|\mathcal{D})$ is analytically intractable due to the high-dimensional integral in the denominator ~\citep{mackay1992practical}. Therefore, \ac{VI} is used to achieve tractable approximate inference ~\citep{graves2011practical, hinton1993keeping}. In this context, \ac{VI} approximates the complex posterior $p(w|\mathcal{D})$ with a simpler variational distribution $q_\theta(w)$, parameterized by $\theta$. The objective is to minimize the Kullback--Leibler (KL) divergence between these distributions. Note that minimizing the KL divergence is mathematically equivalent to maximizing the \ac{ELBO} ~\citep{blundell2015weight}. Consequently, the \ac{ELBO} loss (negative \ac{ELBO}) is minimized during training via \ac{SGD}:

\begin{equation}
    \mathcal{L}_{\text{ELBO}} = -\mathbb{E}_{q_\theta(w)}[\log p(y|x, w)] + \text{KL}[q_\theta(w) \| p(w)]
    \label{eq:elbo_loss}
\end{equation}
Here, the first term $-\mathbb{E}_{q_\theta(w)}[\log p(y|x, w)]$ represents the negative expected log-likelihood (reconstruction error), encouraging data fit. The second term $\text{KL}[q_\theta(w) \| p(w)]$ acts as a regularizer, penalizing deviations of the variational distribution from the prior ~\citep{hinton1993keeping, graves2011practical}.

To ensure computational tractability, the mean-field approximation is adopted, where the weights are modeled with a fully factorized Gaussian distribution~\citep{blundell2015weight}. Under this framework, the variational distribution $q_\theta(w)$ is parameterized by its variational parameters $\mu$ (mean) and $\sigma$ (standard deviation):

\begin{equation}
    q_\theta(w) = \mathcal{N}(w|\mu, \sigma)
    \label{eq:variational_gaussian}
\end{equation}
The parameters $\mu$ and $\sigma$ are learned iteratively by optimizing the \ac{ELBO} loss.

Finally, the predictive distribution $p(y|x, \mathcal{D})$ is approximated through multiple stochastic forward passes on the network using \ac{MC} estimators ~\citep{blundell2015weight, graves2011practical}. Sampling weights $w_s$ from the variational distribution yields:

\begin{equation}
    p(y|x, \mathcal{D}) \approx \frac{1}{S} \sum_{s=1}^{S} p(y|x, w_s), 
    \quad w_s \sim q_\theta(w)
    \label{eq:mc_estimator}
\end{equation}

where $S$ denotes the number of \ac{MC} samples. In practice, $p(y|x, w_s)$ corresponds to the $softmax$ output of the network in the $s$-th forward pass. By averaging these outputs, this method yields a marginalized probability estimate that is robust against overfitting and enables the rigorous quantification of predictive uncertainty ~\citep{lakshminarayanan2017simple}.

\subsection{Theoretical definition of the Confidence--Uncertainty Boundary}
\label{sec:boundary_definition}

In this work, the predictive entropy is adopted as the primary metric for quantifying uncertainty ~\citep{krishnan2020improving}~\citep{leibig2017leveraging}~\citep{kurz2022uncertainty}~\citep{combalia2020uncertainty}. Rooted in information theory, predictive entropy effectively measures the dispersion of the predictive probability distribution, thereby providing a rigorous evaluation of the model's predictive certainty ~\citep{shannon1948mathematical}. All subsequent references to "uncertainty" in this work are denoted by the symbol $U$, which explicitly refers to the predictive entropy of the Bayesian predictive distribution.

\subsubsection{Confidence and uncertainty}
Based on the predictive distribution $p(y|x, \mathcal{D})$ derived in Eq.~(\ref{eq:mc_estimator}), let $\mathbf{p} = [p_1, \dots, p_K]$ denote the predictive probability vector over $K$ classes, where $p_i = p(y=i|x, \mathcal{D})$.

From this vector, two key quantities are derived: the predicted class label and the associated confidence.
The predicted class ($\hat{y}$) is defined as the index corresponding to the maximum probability:
\begin{equation}
    \hat{y} = \operatorname*{argmax}_{i \in \{1,\dots,K\}} p_i
\end{equation}

Consequently, the confidence ($\hat{p}$) is defined as the probability value of the predicted class:
\begin{equation}
    \hat{p} = \max_{i \in \{1,\dots,K\}} p_i = p_{\hat{y}}
    \label{confidence}
\end{equation}
Here, $\hat{p} \in [0, 1]$ serves as a continuous scalar representing the model's certainty regarding its decision $\hat{y}$.

Uncertainty ($U$) is quantified by the predictive entropy of the distribution $\mathbf{p}$ ~\citep{shannon1948mathematical, krishnan2020improving}:
\begin{equation}
    U(\mathbf{p}) = - \sum_{i=1}^{K} p_i \log p_i
    \label{uncertainty}
\end{equation}

\subsubsection{Derivation of uncertainty boundaries}
A key contribution of this work is the exploration of the feasible region of uncertainty for a given confidence level. While confidence and uncertainty are generally inversely correlated, their relationship is mathematically bounded. 
Based on Eq.~\eqref{uncertainty}, the bounds $U_{\text{min}}$ and $U_{\text{max}}$ are identified by considering the behavior of prediction: when the probability mass is concentrated on fewer outcomes, the system is more ordered, and the outcomes are easier to predict, leading to a lower uncertainty. In contrast, when the distribution is more dispersed, the outcomes are harder to predict, resulting in higher entropy.
Consequently, for a fixed confidence $\hat{p}$, the uncertainty $U(\mathbf{p})$ is strictly constrained within a theoretical range $[U_{\text{min}}(\hat{p}), U_{\text{max}}(\hat{p})]$.

\paragraph{Theoretical maximum uncertainty ($U_{\text{max}}$):} 

The uncertainty is maximized when the residual probability mass $(1 - \hat{p})$ is distributed uniformly among the remaining $K-1$ non-target classes. This represents the state of maximum dispersion given a fixed confidence constraint. The upper bound is defined as:
\begin{equation}
    U_{\text{max}}(\hat{p}) = - \left[ \hat{p} \log \hat{p} + \sum_{j \neq \hat{y}} \frac{1 - \hat{p}}{K - 1} \log \left( \frac{1 - \hat{p}}{K - 1} \right) \right]
    \label{eq:u_max}
\end{equation}
where the summation runs over all indices $j$ representing the non-target classes (i.e., $j \in \{1, \dots, K\}$ and $j \neq \hat{y}$). Since the term inside the summation is constant for all $j$, this expression simplifies to:
\begin{equation}
    U_{\text{max}}(\hat{p}) = - \hat{p} \log \hat{p} - (1 - \hat{p}) \log \left( \frac{1 - \hat{p}}{K - 1} \right)
\end{equation}

\paragraph{Theoretical minimum uncertainty ($U_{\text{min}}$):} 

Conversely, the uncertainty is minimized when the residual probability mass $(1 - \hat{p})$ is concentrated entirely into a single non-target class (i.e., the second most likely class), leaving the other $K-2$ classes with zero probability. This represents the most concentrated distribution possible for a fixed $\hat{p}$. The lower bound is defined as:
\begin{equation}
    U_{\text{min}}(\hat{p}) = - \left[ \hat{p} \log \hat{p} + (1 - \hat{p}) \log (1 - \hat{p}) + \sum_{\text{others}} 0 \right]
    \label{eq:u_min}
\end{equation}
which simplifies to:
\begin{equation}
    U_{\text{min}}(\hat{p}) = - \hat{p} \log \hat{p} - (1 - \hat{p}) \log (1 - \hat{p})
\end{equation}

The distinction between $U_{\text{min}}$ and $U_{\text{max}}$ captures qualitatively different modes of predictive uncertainty that cannot be inferred from confidence alone.
When uncertainty approaches $U_{\text{min}}$ for a given $\hat{p}$, the residual probability mass of the model is concentrated on a single alternative class. 
This configuration indicates that the model exhibits a focused conflict between a limited number of alternatives, implying that partial exclusionary knowledge is already present.
From an information-theoretic standpoint, such exclusion reduces the number of plausible outcomes and therefore lowers the uncertainty of the predictive distribution.
Conversely, when the non-dominant predictions are distributed across all remaining classes without preference, the model fails to rule out any alternative hypothesis.
The residual uncertainty is maximally dispersed, indicating a lack of discriminatory evidence.
In this regime, additional information is required to resolve the prediction, leading to a higher uncertainty state.
This configuration corresponds to the maximal achievable uncertainty for the same $\hat{p}$, denoted as $U_{\text{max}}$.

\subsubsection{Confidence--Uncertainty Boundary Curve}

Based on these theoretical bounds, an ideal mapping function is represented, denoted by \ac{CUBC}. A key observation motivating this representation is that in conventional \acp{BNN}, the relationship between confidence and uncertainty spans a continuous region rather than a deterministic correspondence: for any given confidence level, uncertainty can assume a range of values within the feasible bounds. This many-to-one ambiguity limits the ability to provide targeted guidance for uncertainty optimization during training. The \ac{CUBC} addresses this limitation by establishing an explicit functional mapping between confidence and uncertainty, thereby enabling directional optimization that systematically steers the uncertainty distribution toward the desired calibration objectives.

Given a confidence threshold hyperparameter $\gamma$, the ideal uncertainty $U_{\text{ideal}}$ is defined as:
\begin{equation}
    U_{\text{ideal}}(\hat{p}) =
    \begin{cases}
    U_{\text{min}}(\hat{p}) & \text{if } \hat{p} > \gamma \\
    U_{\text{max}}(\hat{p}) & \text{if } \hat{p} \le \gamma
    \end{cases}
    \label{eq:boundary_ideal}
\end{equation}

\begin{figure}[htbp] 
    \centering
    \includegraphics[width=0.6\columnwidth]{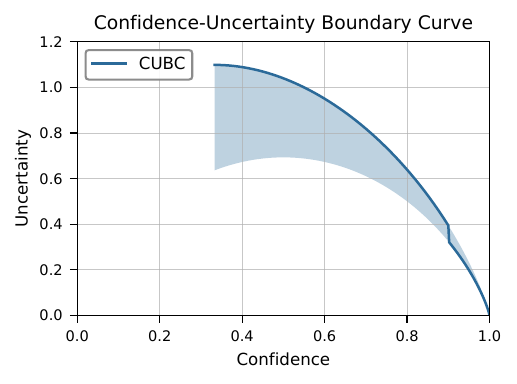} 
    \caption{Visualization of the Confidence--Uncertainty Boundary Curve.}
    \label{fig:boundary_curve}
\end{figure}

The geometric interpretation of this function is visually presented in Figure~\ref{fig:boundary_curve}. The shaded light blue region indicates the feasible mathematical range of uncertainty $[U_{\text{min}}, U_{\text{max}}]$, representing all theoretically attainable confidence--uncertainty pairs. The solid blue line traces the ideal \ac{CUBC} proposed ($U_{\text{ideal}}$) defined in Eq.~(\ref{eq:boundary_ideal}), which collapses this region into a single-valued function that prescribes the target uncertainty for each confidence level.

\subsection{Confidence--Uncertainty Boundary Loss}
\label{sec:cub_loss}

Building upon the \ac{CUBC}, a supervised objective named \ac{CUB-Loss} is proposed. This method encourages the model to exhibit high confidence with low uncertainty for correctly predicted samples and low confidence with high uncertainty for incorrectly predicted samples, thereby achieving more calibrated and dependable model outputs.

\subsubsection{Four-dimensional loss decomposition}

The \ac{AvU} framework~\citep{krishnan2020improving} establishes a fundamental principle for uncertainty calibration by categorizing predictions into four quadrants according to their accuracy and uncertainty characteristics. This paradigm is extended by leveraging the explicit functional mapping provided by the \ac{CUBC}. Rather than relying on soft approximations to partition samples, the proposed formulation directly computes geometric deviations from the ideal boundary curve, yielding a quantitative optimization signal that precisely steers the uncertainty distribution toward calibration targets.

Specifically, the \ac{CUBC} partitions the confidence space into two regimes (high-confidence: $\hat{p} > \gamma$; low-confidence: $\hat{p} \le \gamma$), while prediction correctness naturally divides samples into accurate ($\hat{y}_i = y_i$) and inaccurate ($\hat{y}_i \neq y_i$) categories. The Cartesian product of these two binary dimensions yields four logical regions. Region-specific deviation metrics are designed to be tailored to the optimization objective within each quadrant.

Let $\mathcal{D} = \{(x_i, y_i)\}_{i=1}^N$ denote the dataset, where $N$ is 
the total number of samples. For each sample $i$, we denote $y_i$ as the true 
label, $\hat{y}_i$ as the predicted label, $\hat{p}_i$ as the confidence, and $U_i$ as the uncertainty. The boundary deviation $\delta_i$ is defined with region-dependent calculation logic:

\begin{equation}
    \delta_i =
    \begin{cases}
    | U_{\text{ideal}}(\hat{p}_i) - U_i | & \text{if } \hat{y}_i = y_i, \hat{p}_i > \gamma \\[4pt]
    | \hat{p}_i - \gamma | & \text{if } \hat{y}_i = y_i, \hat{p}_i \le \gamma \\[4pt]
    | \hat{p}_i - \gamma  | & \text{if } \hat{y}_i \neq y_i, \hat{p}_i > \gamma \\[4pt]
    | U_{\text{ideal}}(\hat{p}_i) - U_i | & \text{if } \hat{y}_i \neq y_i, \hat{p}_i \le \gamma
    \end{cases}
    \label{eq:deviation_mixed}
\end{equation}

\noindent where the terms are defined as follows:
\begin{itemize}
    \item $U_{\text{ideal}}(\hat{p}_i)$: the target uncertainty prescribed by the boundary curve for confidence $\hat{p}_i$.
    \item $\gamma$: the confidence threshold that partitions the confidence space.
\end{itemize}

The optimization objective for each region is as follows:
\begin{itemize}
    \item Accurate-Certain ($\hat{y}_i = y_i$, $\hat{p}_i > \gamma$): The prediction is already in the high confidence region with the correct classification. The objective is to further minimize the uncertainty toward $U_{\text{min}}$, reinforcing certain reliable predictions.
    \item Accurate-Uncertain ($\hat{y}_i = y_i$, $\hat{p}_i \le \gamma$): The prediction is correct but remains in the low confidence region. The objective is to increase confidence toward the threshold $\gamma$, calibrating the sample into the high-confidence region.
    \item Inaccurate-Certain ($\hat{y}_i \neq y_i$, $\hat{p}_i > \gamma$): The objective is to decrease confidence toward the threshold $\gamma$, penalizing dangerous overconfident misclassifications.
    \item Inaccurate-Uncertain ($\hat{y}_i \neq y_i$, $\hat{p}_i \le \gamma$): The objective is to increase uncertainty toward $U_{\text{max}}$, encouraging the model to express high uncertainty for incorrect predictions.
\end{itemize}

This four-dimensional decomposition ensures that the optimization signal is geometrically meaningful within each region. Since the \ac{CUBC} establishes an explicit mapping between confidence and uncertainty, it becomes possible to first calibrate confidence to guide predictions into the appropriate region, and subsequently refine uncertainty toward the theoretical bounds. This hierarchical calibration strategy encourages correct predictions to increase confidence while penalizing incorrect predictions that remain in the high-confidence region, systematically guiding all samples toward the ideal \ac{CUBC} illustrated in Figure~\ref{fig:boundary_curve}.

\subsubsection{Normalization and loss formulation}
For the four dimensions of the boundary deviation defined above, a linear normalization is applied, mapping them to a fixed range $[0, 1]$. This process results in the normalized deviation $\tilde{\delta}_i$, thereby ensuring that all four dimensions share a consistent numerical scale in the loss computation.

The final \ac{CUB-Loss} is formulated using a logarithmic barrier function. The loss is aggregated across all samples using a direct summation to impose a heavy penalty as the normalized deviation approaches its maximum:
\begin{equation}
    \mathcal{L}_{\text{CUB}} = \sum_{i=1}^{N} \left( - \log(1 - \tilde{\delta}_i) \right)
    \label{eq:cub_final}
\end{equation}

\subsubsection{Joint optimization strategy}
\label{sec:joint_optimization}

Integrating this regularization into the Bayesian framework, the total objective function combines the \ac{ELBO} with the \ac{CUB-Loss}:
\begin{equation}
    \mathcal{L}_{\text{total}} = \mathcal{L}_{\text{ELBO}} + \beta \cdot \mathcal{L}_{\text{CUB}}
    \label{eq:total_loss}
\end{equation}
where $\mathcal{L}_{\text{ELBO}}$ consists of the cross-entropy loss and the KL divergence term (as defined in Eq.~\ref{eq:elbo_loss}), and $\beta$ is a dynamically adjusted weighting coefficient. A warm-up strategy is adopted, in which $\beta$ is initially set to $0$ during the early training epochs to allow the model to prioritize discriminative learning features. After the warm-up phase, $\beta$ is selected by comparing the magnitude of $\mathcal{L}_{\text{CUB}}$ with the cross-entropy and KL divergence components. Specifically, $\beta$ is calculated such that the weighted term $\beta \cdot \mathcal{L}_{\text{CUB}}$ remains on the same order of magnitude as the cross-entropy and KL divergence losses. This balance strategy ensures that all three loss components contribute comparably to gradient updates, preventing any single objective from dominating the optimization process.

\subsection{Dual Temperature Scaling strategy}
\label{sec:dts_strategy}

Standard \ac{TS} rescales the logits by a single scalar $T > 0$ for all samples~\citep{guo2017calibration}. However, as illustrated in the previous boundary analysis, the ideal relationship between confidence and uncertainty is governed by a piecewise function ($U_{\text{ideal}}$). A single global temperature tends to push the uncertainty of all predictions uniformly toward either the upper or lower boundary, failing to distinguish predictions with high confidence and low uncertainty from those with low confidence and high uncertainty.

\subsubsection{Definition of uncertainty calibration}

To further refine the model's probabilistic outputs and align them with the discovered \ac{CUBC}, a post-hoc calibration method named \ac{DTS} is introduced. \ac{DTS} addresses this limitation by employing two distinct temperature parameters, enabling simultaneous and bidirectional calibration of both confidence and uncertainty toward the ideal boundary curve.

Based on the \ac{CUBC} derived above, the perfect uncertainty calibration is defined as:
\begin{equation}
    U(\mathbf{p}) = U_{\text{ideal}}(\hat{p}), \quad \forall \mathbf{p}
    \label{eq:perfect_calibration}
\end{equation}
where $\hat{p}$ is the confidence in Eq.~(\ref{confidence}), $U(\mathbf{p})$ is the uncertainty in Eq.~(\ref{uncertainty}), and $U_{\text{ideal}}$ is defined in Eq.~(\ref{eq:boundary_ideal}). This condition requires that for any prediction, its uncertainty should equal the ideal value determined by its confidence.

To quantify the degree of calibration, a scalar summary statistic is convenient. Inspired by previous work on uncertainty calibration~\citep{laves2019temperature}, uncertainty miscalibration is defined as the expectation between the predicted uncertainty and the ideal boundary:
\begin{equation}
    \mathbb{E}\left[ \left| U(\mathbf{p}) - U_{\text{ideal}}(\hat{p}) \right| \right]
    \label{eq:uce_expectation}
\end{equation}

\ac{BCCE} approximates Eq.~(\ref{eq:uce_expectation}) by partitioning predictions into $M$ equally-spaced bins based on confidence and taking a weighted average of the bins' uncertainty discrepancy. More precisely:

\begin{equation}
    \text{BCCE} = \sum_{m=1}^{M} \frac{|B_m|}{N} \big| \bar{U}(B_m) - \bar{U}_{\text{ideal}}(B_m) \big|
    \label{eq:bcce_main}
\end{equation}

where $M$ is the number of bins, and $B_m$ denotes the set of indices for samples falling into the $m$-th bin. The terms $\bar{U}$ and $\bar{U}_{\text{ideal}}$ are defined as:

\begin{equation}
    \bar{U}(B_m) = \frac{1}{|B_m|} \sum_{i \in B_m} U(\mathbf{p}_i) \quad 
    \label{eq:bcce_real_uncertainty}
\end{equation}

\begin{equation}
   \quad \bar{U}_{\text{ideal}}(B_m) = \frac{1}{|B_m|} \sum_{i \in B_m} U_{\text{ideal}}(\hat{p}_i)
    \label{eq:bcce_target_uncertainty}
\end{equation}

Here, $U(\mathbf{p}_i)$ represents the actual uncertainty calculated from the model's output distribution for sample $i$, and $U_{\text{ideal}}(\hat{p}_i)$ represents the target uncertainty value derived from the \ac{CUBC} corresponding to the sample's confidence $\hat{p}_i$. The discrepancy between the model's actual uncertainty and the theoretically required uncertainty across the entire confidence spectrum is captured by this formulation.

\subsubsection{The Dual Temperature Scaling algorithm}
\label{sec:dts_algorithm}

The \ac{DTS} algorithm acts as a bi-directional optimization process characterized by two core properties:
\begin{itemize}
\item \textit{Piecewise optimization:} It enables independent optimization of samples in high-confidence and low-confidence regimes, effectively approximating the respective segments of the \ac{CUBC}.
\item \textit{Accuracy and \acs{AvU} preservation:} Positive temperature scaling preserves the logit ranking within each stochastic forward pass and therefore mainly changes the confidence and uncertainty profile rather than the learned decision structure. In the Bayesian setting, probability averaging across stochastic forward passes does not imply strict ranking preservation after aggregation; nevertheless, with frozen model weights and fixed pre-scaling region assignments, \acs{DTS} is expected to preserve the reported accuracy in the evaluated settings. The \acs{AvU} score changes only when calibrated confidence or uncertainty values cross the operating thresholds that define the accuracy and uncertainty categories. Because \acs{DTS} assigns scaling regions before temperature adjustment, most samples remain on the same side of these thresholds. This allows \acs{DTS} to reduce \acs{BCCE} by moving the continuous confidence and uncertainty profile closer to the \ac{CUBC}, while the reported \acs{AvU} exhibits only negligible variation.
\end{itemize}

\begin{figure}[ht]
    \centering
    \includegraphics[width=0.6\linewidth]{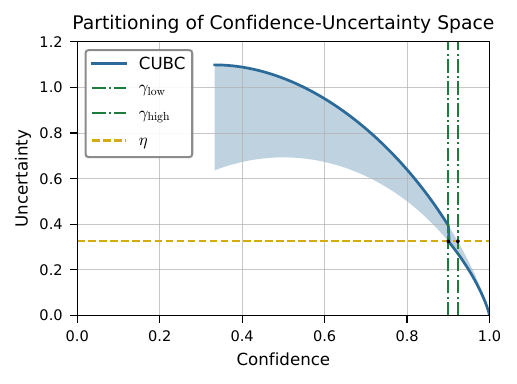}
    \caption{Partitioning of the confidence--uncertainty space by thresholds $\gamma_{\text{low}}$, $\gamma_{\text{high}}$, and $\eta$.}
    \label{fig:dts_mechanism}
\end{figure}

A confidence threshold interval $[\gamma_{\text{low}}, \gamma_{\text{high}}]$ and an uncertainty threshold $\eta$ are defined based on the \ac{CUBC}: given $\eta$, $\gamma_{\text{low}}$ and $\gamma_{\text{high}}$ are the confidence values at which $U_{\text{min}}$ and $U_{\text{max}}$ equal $\eta$, respectively. These three thresholds are mutually determined. The schematic representation of these thresholds and the resulting scaling regions is illustrated in Figure~\ref{fig:dts_mechanism}.

To enhance algorithmic clarity and probabilistic confidence level, the scaling mechanism is structured into three sequential steps:

\vspace{0.5em}
\noindent \textit{Step 1: Distribution sharpening ($T_{\text{high}}$)} \\
First, samples that demonstrate high reliability are identified. The temperature $T_{\text{high}}$ is applied to sharpen the predictive distribution (reducing $U(\mathbf{p})$ towards $U_{\text{min}}$) when either of the following conditions is satisfied:
\begin{itemize}
    \item The prediction is high-confidence ($\hat{p} > \gamma_{\text{high}}$).
    \item The prediction falls into the intermediate confidence range with low uncertainty ($\gamma_{\text{low}} < \hat{p} \le \gamma_{\text{high}}$ and $U(\mathbf{p}) < \eta$).
\end{itemize}

\vspace{0.5em}
\noindent \textit{Step 2: Distribution softening ($T_{\text{low}}$)} \\
Second, samples that require preserved ambiguity are identified. The temperature $T_{\text{low}}$ is applied to soften the predictive distribution (increasing $U(\mathbf{p})$ towards $U_{\text{max}}$) when either of the following conditions is satisfied:
\begin{itemize}
    \item The prediction is low-confidence ($\hat{p} \le \gamma_{\text{low}}$).
    \item The prediction falls into the intermediate confidence range with high uncertainty ($\gamma_{\text{low}} < \hat{p} \le \gamma_{\text{high}}$ and $U(\mathbf{p}) \ge \eta$).
\end{itemize}

\vspace{0.5em}
\noindent \textit{Step 3: Bi-directional Temperature Scaling} \\
Given the conditions established in Steps 1 and 2, post-hoc calibration for the VI-based \ac{BNN} is applied to each stochastic logit vector before Bayesian predictive aggregation. Let $\mathbf{z}_s$ denote the logit vector from the $s$-th \ac{MC} forward pass. The region-specific scaling operation is applied to each $\mathbf{z}_s$ individually:
\begin{equation}
{
    \mathbf{z}_{s,\text{calibrated}} =
    \begin{cases}
    \mathbf{z}_s / T_{\text{high}} & \text{if Step~1 conditions are satisfied} \\
    \mathbf{z}_s / T_{\text{low}}  & \text{if Step~2 conditions are satisfied}
    \end{cases}
}
    \label{eq:dual_temperature}
\end{equation}
The calibrated predictive distribution is then obtained by averaging the corresponding $softmax$ probabilities across all \ac{MC} forward passes:
\begin{equation}
{
    \mathbf{p}_{\text{calibrated}} =
    \frac{1}{S}\sum_{s=1}^{S}
    \operatorname{softmax}\!\left(\mathbf{z}_{s,\text{calibrated}}\right)
}
    \label{eq:mc_aggregation}
\end{equation}
The intermediate confidence region $(\gamma_{\text{low}}, \gamma_{\text{high}})$ combined with the uncertainty threshold $\eta$ introduces a transition zone that prevents abrupt scaling changes. This design enables the model to effectively distinguish between ``confident-accurate'' and ``uncertain-inaccurate'' predictions during the calibration process.

\subsubsection{Optimization of temperature parameters}
Following the standard validation protocol for temperature scaling~\citep{guo2017calibration}, the temperature parameters $\mathcal{T} = \{T_{\text{high}}, T_{\text{low}}\}$ were selected on the validation split for each evaluation dataset. The optimization procedure consists of three sequential steps: 1) assigning validation samples to fixed \acs{DTS} regions according to the unscaled Bayesian predictive distribution; 2) applying candidate temperature pairs via the per-\ac{MC} scaling and Bayesian $softmax$ aggregation defined in Eqs.~(\ref{eq:dual_temperature})--(\ref{eq:mc_aggregation}); and, 3) selecting the temperature pair that minimizes the validation-set \acs{BCCE} with respect to the proposed operational boundary target:
\begin{equation}
    \mathcal{T}^* = \operatorname*{argmin}_{T_{\text{high}},
    T_{\text{low}}} \; \text{BCCE}(\mathcal{D}_{\text{val}};
    T_{\text{high}}, T_{\text{low}})
\end{equation}
With the model weights frozen throughout, only the two temperature parameters are varied during calibration. The resulting temperature pair $\mathcal{T}^*$ is subsequently applied once to the corresponding held-out test set using the same pre-scaling region assignment rule.
The resulting model exhibits a confidence--uncertainty relationship that closely approximates the ideal boundary curve, improving predictive reliability for clinical deployment.

\section{Experiments}\label{sec:experiments}

The proposed framework is evaluated retrospectively on three medical imaging tasks: screening of pneumonia types, detection of \ac{DR}, and identification of skin lesions. The pneumonia screening task serves as the primary setting for evaluating the full calibration pipeline, while the \ac{DR} detection task evaluates robustness under reduced-data conditions, and the skin lesion identification task tests applicability under severe class imbalance, collectively demonstrating the generalizability of the framework across diverse imaging modalities, class distributions, and data regimes.

A subsection is dedicated to each of the application domains (i.e., experiments) used to evaluate the proposed methods.

\subsection{Experimental setup and evaluation}

The screening of pneumonia types serves as the primary experimental testbed, where a four-stage analysis is performed to evaluate the main components of the proposed framework.

\begin{enumerate}
    \item First, a \ac{VI}-based \acp{BNN} is established as a baseline to represent standard probabilistic modeling.
    \item Second, the proposed learning with \ac{CUB-Loss} is integrated into the Bayesian training process to assess its impact on intrinsic calibration. 
    \item Third, the post-hoc \ac{DTS} mechanism, optimized via the \ac{BCCE} metric, is applied to the \ac{CUB-Loss}-trained model to demonstrate the benefits of the complete calibration pipeline. 
    \item Finally, the robustness of the model is evaluated by testing with the \ac{OOD} dataset to ensure reliability in open-world settings.
\end{enumerate}

For the \ac{DR} detection and skin lesion identification datasets, a streamlined protocol following the first three evaluation stages is adopted, assessing the baseline Bayesian model, the contribution of the \ac{CUB-Loss} to the training process, and the subsequent \ac{DTS} post-hoc calibration. This design allows consistent performance gains to be demonstrated across different clinical domains without redundant ablation steps.

\subsubsection{Evaluation metrics}

In accordance with the first three evaluation stages, different metrics are employed for each comparative analysis. When comparing the baseline Bayesian model and the \ac{CUB-Loss}-trained model (Stages~1 to~2), the classification accuracy, and the \ac{AvU} metric~\citep{krishnan2020improving} are employed to assess the impact of training-time on predictive correctness and the alignment between model confidence and reliability. When comparing the \ac{CUB-Loss}-trained model and the \ac{DTS}-calibrated configuration (Stages~2 to~3), \ac{BCCE} serves as the primary metric, as it directly quantifies deviation from the target \ac{CUBC} boundary. In this last scenario, \ac{ECE} and \ac{UCE} are reported as complementary indicators for supplementary calibration analysis.

For the \ac{OOD} detection task (stage 4), the problem is modeled as a binary classification task that distinguishes between \ac{ID} and \ac{OOD} samples based on their predictive uncertainty. Following the evaluation protocol established in work~\citep{hendrycks2017baseline}, two widely used metrics are employed: the \ac{AUROC} and the \ac{AUPR}. \ac{AUROC} measures the probability that a randomly chosen \ac{OOD} sample receives a higher anomaly score than a randomly chosen \ac{ID} sample, while \ac{AUPR} summarizes the precision-recall trade-off across varying thresholds. Higher values in both metrics indicate that the model assigns consistently higher uncertainty scores to \ac{OOD} samples compared to \ac{ID} samples, effectively flagging them as ``unknown'' to prevent silent failures.

\subsection{Experiment 1: screening of pneumonia types from chest X-ray images}

Chest radiography is a vital complementary modality for the screening of pneumonia and the detection of its specific etiology: viral, bacterial, or fungal~\citep{cleverley2020role}. Although deep learning approaches have demonstrated remarkable potential in automating the analysis of X-ray images for this purpose~\citep{ozturk2020automated}, a potential clinical deployment requires more than just high classification accuracy in known categories.

\subsubsection{Datasets }
\label{sec:dataset}

Beyond accurate classification of known categories, a clinically reliable screening system must also identify inputs that fall outside its learned distribution~\citep{gonzalezDistancebasedDetectionOutofdistribution2022}. This capability is essential for detecting ``silent failures'' where the model might otherwise produce overconfident but erroneous predictions on unfamiliar pathologies. 

Thus, classification performance is evaluated on \ac{ID} data, and reliability is subsequently assessed by rejecting samples from an \ac{OOD} dataset.

\paragraph{In-Distribution dataset:}

The \ac{ID} dataset was compiled by aggregating data from multiple publicly available repositories to create a large-scale corpus. This multi-source construction was motivated by two practical constraints: no single publicly available corpus of chest radiographs covers all three target classes with sufficient sample volume, and individual public datasets are frequently characterized by class imbalance. Because class construction draws on multiple public repositories with distinct acquisition protocols, possible source-related acquisition and preprocessing effects cannot be fully excluded, and complete cross-source patient-level deduplication cannot be verified from public metadata. Accordingly, this dataset is intended as a retrospective evaluation corpus for validating the proposed uncertainty calibration pipeline, and the results are interpreted as evidence of uncertainty-alignment behavior under heterogeneous acquisition conditions rather than as a claim of diagnostic model performance. The dataset comprises over 56,000 posterior-anterior and anterior-posterior chest radiographs, categorized into three classes: Pneumonia (non-COVID-19 viral, bacterial, or fungal), COVID-19, and Control (no findings). The corpus was randomly partitioned into training, validation, and test subsets with a ratio of 70:15:15, using stratified splitting to preserve the class distribution. The detailed distribution is presented in Table~\ref{tab:dataset_distribution}.

\begin{table}[htbp]
\centering
\caption{Distribution of In-Distribution datasets.}
\label{tab:dataset_distribution}
\begin{tabular}{lcccc}
\toprule
Subset & Pneumonia & COVID-19 & Control & Total \\
\midrule
Train       & 10,676 & 13,999 & 14,905 & 39,580 \\
Validation  & 2,287  & 2,999  & 3,193  & 8,479 \\
Test        & 2,290  & 2,999  & 3,194  & 8,483 \\
\midrule
Total & 15,253 & 19,997 & 21,292 & 56,542 \\
\bottomrule
\end{tabular}
\end{table}

The pneumonia class was compiled by combining samples from MIMIC-CXR~\citep{johnson2019mimic} and CheXpert~\citep{irvin2019chexpert} to enhance intra-class variability and introduce realistic clinical complexity.

The COVID-19 class was curated from the COVIDx CXR-4 dataset~\citep{Wang2020}, which consolidates imaging data from multiple repositories including the COVID-19 Chest X-ray dataset~\citep{covid-chestxray}, Actualmed COVID-19 Chest X-ray dataset~\citep{actualmed-covid-chestxray}, COVID-19 Radiography database~\citep{covid-radiography-database-v3}, RSNA International COVID-19 Open Radiology database~\citep{rsna-pneumonia-detection}, BIMCV-COVID19+~\citep{bimcv-covid19-plus}, and Stony Brook University COVID-19 Positive Cases~\citep{covid19-ny-sbu}. From this consolidated corpus, approximately 20,000 COVID-19 positive images were selected.

The control class was sourced from the NIH ChestX-ray14 dataset~\citep{wang2017chestxray}, strictly selecting images labeled as ``No Finding'' to minimize label noise and clinical ambiguity.

\paragraph{Out-of-Distribution dataset:}

An external near-\ac{OOD} stress test was constructed from the NIH ChestX-ray14 dataset~\citep{wang2017chestxray} to evaluate the framework's response to distributional shifts. Unlike standard Far-\ac{OOD} benchmarks that utilize semantically distinct natural images, this evaluation focuses on non-target thoracic findings within the same chest radiography domain.

This setup presents high complexity because the \ac{OOD} samples share similar anatomical structure and grayscale radiographic appearance with the \ac{ID} data. Five pathological categories (Pneumothorax, Emphysema, Fibrosis, Nodule/Mass, and Atelectasis) --none present in the \ac{ID} label set-- were selected based on sufficient sample availability. The detailed composition of the \ac{OOD} dataset is detailed in Table~\ref{tab:ood_composition}.

\begin{table}[htbp]
\centering
\caption{Composition of the Near-\ac{OOD} test set.}
\label{tab:ood_composition}
\begin{tabular}{lc}
\toprule
Pathology Category & Sample Count \\
\midrule
Pneumothorax    & 1,457 \\
Emphysema       & 1,457 \\
Fibrosis        & 1,215 \\
Nodule / Mass   & 2,914 \\
Atelectasis     & 1,457 \\
\midrule
Total \ac{OOD} Samples & 8,500 \\
\bottomrule
\end{tabular}
\end{table}

\subsubsection{Model architecture and training}
\label{sec:method_network}

The foundation of the screening system developed is the COVID-Net-Large architecture proposed in~\citep{Wang2020}. This specialized architecture was selected because it utilizes a lightweight projection-expansion-projection-extension design pattern explicitly tailored for chest radiography. This domain-specific design offers superior feature extraction capabilities for thoracic pathology compared to general-purpose models adapted via transfer learning.

Thes COVID-Net-Large architecture was adapted to create a Bayesian version: BCXR-Net.  

\paragraph{BCXR-Net:}
 
To adapt the architecture, the classification head was first modified, which consists of three fully connected layers. Dropout regularization with a rate of 0.2 was incorporated after the first two fully connected layers (before the final classifier layer) to provide complementary regularization for improved generalization. The modified network was then converted into a Bayesian variant, denoted as BCXR-Net, by replacing standard convolutional and linear layers with their Bayesian counterparts using the reparameterization trick~\citep{blundell2015weight, krishnan2022bayesiantorch}. By treating the weights as random variables sampled from learned distributions rather than point estimates, this architecture enables the direct integration of uncertainty quantification into the forward pass~\citep{arias2024analysis}.

\paragraph{Integration of Confidence--Uncertainty Boundary Loss:}
To further enhance the discriminative capability of the model's uncertainty estimates, the proposed \ac{CUB-Loss} is integrated into the Bayesian framework. The total objective function defined in Eq.~(\ref{eq:total_loss}) is optimized. Following the joint optimization strategy described in Section~\ref{sec:joint_optimization}, a warm-up strategy is employed in which $\beta$ is set to 0 for the first 5 epochs and then fixed at 0.1 for the remaining training epochs. For the specific configuration of the \ac{CUB-Loss}, the confidence threshold is set to $\gamma = 0.9$, which corresponds to a theoretical uncertainty lower bound of approximately 0.325.

\paragraph{Training protocol and hyperparameters:}
The experiment was implemented using the PyTorch\textsuperscript{®} framework. All experiments were conducted with a fixed random seed of 42 to ensure reproducibility.

For preprocessing, input images were resized to 256 pixels on the shorter side and then center-cropped to $240 \times 240$ pixels. Although the original chest radiographs are grayscale, the images were processed as 3-channel RGB input by replicating the single intensity channel. This strategy maintains architectural compatibility with backbones pre-trained on general-purpose color datasets, thereby preserving the flexibility to leverage transfer learning. All images were normalized using standard ImageNet statistics (mean = [0.485, 0.456, 0.406], std = [0.229, 0.224, 0.225]) for each channel. Data augmentation was applied during training to improve generalization, including random resized cropping (scale range from 0.5 to 1.0) and random horizontal flipping.

The network was trained using the \ac{SGD} optimizer with a momentum of 0.9 and a weight decay of $5 \times 10^{-4}$. A cosine annealing learning rate schedule was employed to smoothly decay the learning rate from an initial value of $4 \times 10^{-4}$ over 50 epochs with a batch size of 128.

Following prior uncertainty-calibration practice, predictive probabilities and entropy-based uncertainties are estimated by averaging multiple stochastic forward passes; the \ac{AvUC} implementation employs 128 inference-time \ac{MC} samples~\citep{krishnan2020improving}, indicating that relatively large sampling counts are commonly adopted to stabilize uncertainty estimates. Based on the experimental experience of this work, 80 \ac{MC} forward passes were selected for inference as a practical balance between uncertainty-estimation stability and computational cost. For the stochastic variational layers, 5 \ac{MC} samples per forward pass were used during training to balance computational efficiency with gradient estimation accuracy. The same \ac{MC} sampling settings are adopted as the hyperparameter configuration across all datasets evaluated in this work.

\paragraph{Post-hoc calibration protocol:}
After training, the \ac{DTS} mechanism described in Section~\ref{sec:dts_strategy} was applied to recalibrate the model's probabilistic outputs. The calibration was performed on the held-out validation set with frozen model weights to prevent data leakage.

Following the threshold derivation described in Section~\ref{sec:dts_algorithm}, the uncertainty threshold was set to $\eta = 0.325$, and the corresponding confidence thresholds $\gamma_{\text{low}} = 0.9$ and $\gamma_{\text{high}} = 0.923$ were derived from the \ac{CUBC} at this uncertainty level. The optimal temperature parameters $\mathcal{T} = \{T_{\text{high}}, T_{\text{low}}\}$ were determined by minimizing the \ac{BCCE} with fixed model weights in the validation-set. The same threshold configuration is applied consistently across all datasets evaluated in this work.

\subsubsection{Experimental results}

The results are presented separately for the \ac{ID} and \ac{OOD} datasets. 

\paragraph{In-Distribution Training-Time Calibration:}

\ac{MC} Dropout provides a practical stochastic approximation to posterior inference by interpreting inference-time dropout as a lightweight variational Bayesian approximation~\citep{gal2016dropout}, and is widely used as a baseline to evaluate the stochastic uncertainty~\citep{laves2019temperature,ovadia2019can}. It is therefore included here as a representative stochastic baseline. The \ac{MC} Dropout experimental configuration follows the protocol of~\citep{laves2019temperature}.

To evaluate the training-time contribution of the proposed objective, Table~\ref{tab:results} compares BCXR-Net, CXR-Net with \ac{MC} Dropout (where CXR-Net is the deterministic model counterpart to BCXR-Net), BCXR-Net with $\mathcal{L}_{\text{AvUC}}$~\citep{krishnan2020improving}, and BCXR-Net with the proposed $\mathcal{L}_{\text{CUB}}$.  To quantify the training-time calibration effectiveness, the uncertainty separation metric $\Delta U = \bar{U}_{\text{incorrect}} - \bar{U}_{\text{correct}}$ is introduced, which measures the difference in mean uncertainty between incorrect and correct predictions. A higher $\Delta U$ indicates that the model assigns high uncertainty to erroneous predictions and low uncertainty to correct ones, thereby reflecting better calibration quality. Fig.~\ref{fig:cxr_uncertainty_kde_overlay} shows the KDE curves of predictive uncertainty for correct and incorrect predictions under each configuration.

\begin{table}[htbp]
\centering
\caption{Training-time results on the \ac{ID} dataset.}
\label{tab:results}
\begin{tabular}{lccc}
\toprule
Method & Acc. (\%) $\uparrow$ & AvU $\uparrow$ & $\Delta U$ $\uparrow$ \\
\midrule
BCXR-Net & 95 & 0.43 & 0.26 \\
CXR-Net + MC Dropout & 93 & 0.63 & 0.34 \\
BCXR-Net + $\mathcal{L}_{\text{AvUC}}$ & 84 & 0.59 & 0.32 \\
BCXR-Net + $\mathcal{L}_{\text{CUB}}$ & \textbf{97} & \textbf{0.90} & \textbf{0.45} \\
\bottomrule
\end{tabular}

\vspace{0.55em}
\includegraphics[width=0.6\linewidth]{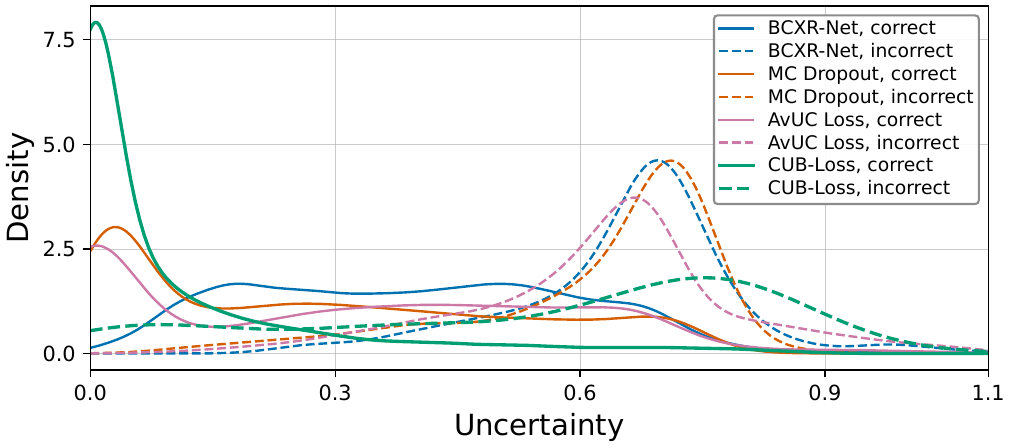}
\captionof{figure}{Uncertainty KDEs for correct and incorrect predictions
across the methods presented in Table \ref{tab:results}.}
\label{fig:cxr_uncertainty_kde_overlay}
\end{table}

The baseline Bayesian model achieves an accuracy of 95\% with an \ac{AvU} of 0.43. As shown in Fig.~\ref{fig:cxr_uncertainty_kde_overlay}, the KDE curves of its correct and incorrect predictions span a broad uncertainty range and exhibit substantial overlap across the mid-uncertainty region, with no clear separation between the two distributions. This indicates that, while \ac{VI}-based posterior inference captures the uncertainty of the parameter, it does not by itself align uncertainty with prediction correctness.

CXR-Net with \ac{MC} Dropout increases \ac{AvU} from 0.43 to 0.63 and $\Delta U$ from 0.26 to 0.34, showing improved alignment between uncertainty and prediction correctness relative to the Bayesian baseline. However, this improvement is accompanied by lower classification accuracy (93\% vs. 95\%). In any case, these results must be analised in view of the results presented in \citep{djupskaas2026unreliable}, that cautions against treating \ac{MC} Dropout uncertainty as a fully reliable epistemic posterior. This limitation motivates evaluating whether a \ac{VI}-based Bayesian backbone can better satisfy the \ac{AvU} framework objective.

For $\mathcal{L}_{\text{AvUC}}$, the \ac{AvU} score increases from 0.43 to 0.59, and $\Delta U$ increases from 0.26 to 0.32 relative to the plain Bayesian baseline. However, these values remain lower than those using \ac{MC} Dropout (\ac{AvU} 0.63 and $\Delta U$ 0.34), and the accuracy decreases to 84\%. This behavior is attributed to the design of $\mathcal{L}_{\text{AvUC}}$, whose optimization is based on the sample counts $n_{\mathrm{AC}}$, $n_{\mathrm{AU}}$, $n_{\mathrm{IC}}$, and $n_{\mathrm{IU}}$. This count-based structure may improve aggregate \ac{AvU}, while offering only indirect control over how individual predictions approach the target boundary.

By contrast, $\mathcal{L}_{\text{CUB}}$ penalizes deviations from the boundary target at the individual prediction level, providing a direct supervision signal through the \ac{CUBC} boundary target. With this objective, BCXR-Net achieves the best overall performance, reaching an accuracy of 97\%, an \ac{AvU} of 0.90, and $\Delta U$ of 0.45. As shown in Fig.~\ref{fig:cxr_uncertainty_kde_overlay}, the correct-prediction KDE is concentrated near low uncertainty, while the incorrect-prediction KDE is broadly shifted towards high uncertainty values, yielding a clearer separation than the other configurations. Overall, these results suggest that \ac{CUB-Loss} provides a more effective calibration objective, improving uncertainty alignment and yielding better results than \ac{MC} Dropout and the Bayesian+$\mathcal{L}_{\text{AvUC}}$ baseline.

\paragraph{In-Distribution Post-hoc Calibration:}

Table~\ref{tab:posthoc_calibration} reports the test-set calibration metrics, and Fig.~\ref{fig:bcce_comparison} visualizes the corresponding empirical confidence--uncertainty profiles relative to the \ac{CUBC} boundary target. Three post-hoc strategies are compared: 1) \acs{TS}-NLL as the standard probability-calibration baseline; 2) \acs{TS}-\acs{BCCE} to isolate the effect of the scalar selection objective; and, 3) \acs{DTS}-\acs{BCCE} to assess whether region-specific scaling improves boundary alignment.

\begin{table*}[htbp]
\centering
\caption{Post-hoc calibration comparison on the \ac{ID} test set.}
\label{tab:posthoc_calibration}
\footnotesize
\setlength{\tabcolsep}{6pt}
\begin{tabular}{lcccccc}
\toprule
Method & Acc. (\%) $\uparrow$ & ECE(\%) & UCE (\%) & BCCE (\%) $\downarrow$ & AvU $\uparrow$ & $\Delta U$ $\uparrow$ \\
\midrule
$\mathcal{L}_{\text{CUB}}$ & 97 & 0.56 & 7.08 & 1.44 & 0.90 & 0.45 \\
$\mathcal{L}_{\text{CUB}}$ + \acs{TS}-NLL & 97 & 0.39 & 6.40 & 1.45 & 0.90 & 0.43 \\
$\mathcal{L}_{\text{CUB}}$ + \acs{TS}-\acs{BCCE} & 97 & 0.47 & 5.75 & 1.46 & 0.91 & 0.41 \\
$\mathcal{L}_{\text{CUB}}$ + \acs{DTS}-\acs{BCCE} & 97 & 1.00 & 6.34 & 0.79 & 0.90 & 0.51 \\
\bottomrule
\end{tabular}
\end{table*}

\begin{table}[htbp]
\centering
\vspace{0.55em}
\includegraphics[width=0.6\linewidth]{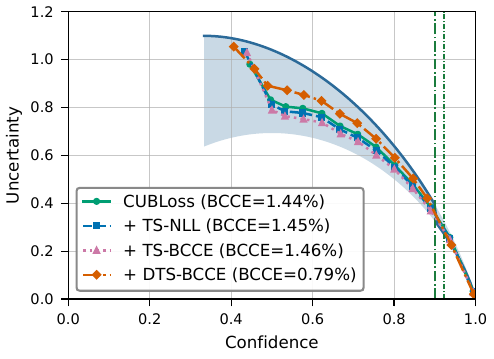}
\captionof{figure}{Reliability curves for calibration 
toward the boundary target on the \ac{ID} test set.}
\label{fig:bcce_comparison}
\end{table}

The proposed \ac{CUBC} defines the target relation that the model is expected to satisfy between confidence and uncertainty. During training, $\mathcal{L}_{\text{CUB}}$ penalizes predictions that deviate from this boundary, encouraging correct predictions to carry lower uncertainty and incorrect predictions to carry higher uncertainty. After training, \ac{DTS} uses the same boundary target to refine the relation between confidence and uncertainty without updating model weights.

To quantify this boundary alignment, \acs{BCCE} is adopted as the primary boundary-alignment metric, while conventional \acs{UCE} is retained as a complementary uncertainty-calibration diagnostic. \acs{UCE} measures calibration through a bin-wise identity relation between predictive uncertainty and empirical error rate, implicitly assuming that the desired correspondence between uncertainty and error is linear. In contrast, the \ac{CUBC} boundary is curved in the space of confidence and uncertainty. Since the empirical error rate is expected to vary with confidence, the relationship between uncertainty and error arising from boundary alignment is not necessarily linear. Therefore, \acs{BCCE} is structurally consistent with the proposed framework, as it directly measures deviation of the empirical confidence and uncertainty profile from the curved \ac{CUBC} target.

Table~\ref{tab:posthoc_calibration} and Fig.~\ref{fig:bcce_comparison} demonstrate that the three post-hoc strategies produce distinct calibration outcomes. The two scalar baselines show a consistent directional pattern. Although \acs{TS}-NLL and scalar \acs{TS}-\acs{BCCE} optimize different objectives, both apply a single global temperature and produce the same metric direction: \acs{ECE} and \acs{UCE} decrease relative to the uncalibrated \acs{CUB-Loss} model, while \acs{BCCE} increases respectively. Scalar \acs{TS}-\acs{BCCE} provides the lowest \acs{UCE} value (5.75\%), indicating that a \acs{BCCE}-oriented scalar objective can improve the reported uncertainty-error consistency diagnostic. However, this improvement does not translate into better boundary alignment. As shown in Fig.~\ref{fig:bcce_comparison}, a single global temperature can soften or sharpen logits but has limited ability to reshape the confidence--uncertainty profile toward the curved \ac{CUBC} boundary.

\acs{DTS}-\acs{BCCE} produces a different pattern: \acs{BCCE} decreases from 1.44\% to 0.79\%, while \acs{UCE} decreases to 6.34\%, \acs{ECE} increases to 1.00\%, and $\Delta U$ increases from 0.45 to 0.51. This larger uncertainty gap between incorrect and correct predictions is consistent with \acs{DTS} calibration. Unlike single-temperature scaling, which applies a uniform rescaling that tends to move all calibration metrics in the same direction, \acs{DTS} optimizes separate temperatures for the two \ac{CUBC} boundary regions independently. Each temperature is selected to minimize \acs{BCCE}, not to reduce \acs{ECE} or \acs{UCE}. When the boundary-aligned adjustment directions in both regions do not match the direction that reduces \acs{ECE} or \acs{UCE}, such divergence is a mechanistically expected consequence of optimizing a structurally distinct objective. Therefore, the metric divergence reflects that \acs{DTS}-\acs{BCCE} optimizes for boundary alignment with the curved \ac{CUBC} target, which is a structurally distinct objective from global confidence calibration.

\paragraph{Out-of-Distribution detection:}

\ac{OOD} detection is formulated as a binary classification problem using a scoring function that assigns each sample a value indicating its likelihood of being \ac{ID} or \ac{OOD}~\citep{hendrycks2017baseline}. Unlike standard Far-\ac{OOD} benchmarks that utilize semantically distinct natural images~\citep{zhangOpenOODV15Enhanced2024}, the Near-\ac{OOD} evaluation presents a significantly more challenging scenario.

Based on the BCXR-Net model, the proposed method was compared with several established \ac{OOD} detection approaches. For output-based methods, the maximum $softmax$ probability (confidence) was employed as the scoring function~\citep{devriesLearningConfidenceOutofDistribution2018}; in contrast, uncertainty was utilized as the primary indicator in the proposed method~\citep{macedoEntropicOutofDistributionDetection2022}. Regarding feature-based methods, the Mahalanobis distance and Relative Mahalanobis distance with single-layer and multi-layer feature fusion were evaluated~\citep{anthonyUseMahalanobisDistance2023}. Additionally, Relative Mahalanobis combined with input perturbation was examined~\citep{liangEnhancingReliabilityOutofdistribution2018}. Kernel \ac{PCA} variants that exploit non-linear mappings were also evaluated~\citep{fangKernelPCAOutofdistribution2024}: in Cosine + \ac{PCA}, cosine normalization is applied prior to dimensionality reduction, whereas in Cosine + \ac{RFF} + \ac{PCA}, \ac{RFF} is employed to approximate Gaussian kernels for enhanced \ac{OOD} separability.

\begin{table}[htbp]
\centering
\caption{Results on the Near-\ac{OOD} test set.}
\label{tab:ood_results}
\begin{tabular}{lcc}
\toprule
Method & AUROC $\uparrow$ & AUPR $\uparrow$ \\
\midrule
\multicolumn{3}{l}{\textit{Output-based methods}} \\
$\mathcal{L}_{\text{CUB}}$ (confidence) & 71.1 & 73.7 \\
$\mathcal{L}_{\text{CUB}}$ (uncertainty) & 71.1 & 73.7 \\
$\mathcal{L}_{\text{CUB}}$ + DTS (confidence) & 72.2 & 76.6 \\
$\mathcal{L}_{\text{CUB}}$ + DTS (uncertainty) & \textbf{74.5} & \textbf{79.6} \\
\midrule
\multicolumn{3}{l}{\textit{Feature-based methods}} \\
Mahalanobis & 61.2 & 68.9 \\
Relative Mahalanobis (single-layer) & 62.3 & 69.4 \\
Relative Mahalanobis (multi-layer) & 64.0 & 71.9 \\
Relative Mahalanobis + perturbation & 63.9 & 71.8 \\
Cosine + PCA & 59.0 & 68.6 \\
Cosine + RFF + PCA & 65.0 & 72.6 \\
\bottomrule
\end{tabular}
\end{table}

Table~\ref{tab:ood_results} presents the \ac{OOD} detection results. The moderate overall performance across all methods reflects the inherent difficulty of Near-\ac{OOD} detection in medical imaging~\citep{zhangOpenOODV15Enhanced2024, anthonyUseMahalanobisDistance2023}, where discriminative lesion regions occupy only a small portion of the image area~\citep{gonzalezDistancebasedDetectionOutofdistribution2022}. The proposed calibrated uncertainty (BCXR-Net + $\mathcal{L}_{\text{CUB}}$ + \ac{DTS}) achieves the best performance with an \ac{AUROC} of 74.5\% and \ac{AUPR} of 79.6\%, outperforming both output-based baselines and feature-based alternatives. These results demonstrate that the proposed framework provides meaningful \ac{OOD} detection capability even in the challenging Near-\ac{OOD} regime, enhancing both \ac{ID} classification reliability and the identification of out-of-scope inputs.

\subsection{Experiment 2: diabetic retinopathy detection}

\ac{DR} represents a critical microvascular sequela of diabetes mellitus and persists as a leading cause of preventable blindness among the working-age population globally.
Although fundus photography constitutes the primary detection modality, manual grading relies heavily on subjective expert interpretation.

While widespread research has explored deep learning algorithms for automating this problem, annotated fundus data remain limited in scale and severity distributions, being also imbalanced across grading levels. Thus, this application domain poses challenges for uncertainty calibration under reduced-data and class-imbalance conditions. The \ac{DR} experiment is therefore used here as a complementary cross-domain evaluation setting to examine whether the proposed pipeline produces more informative uncertainty behavior under reduced-data and class-imbalance conditions~\citep{akramUncertaintyawareDiabeticRetinopathy2025}.

\subsubsection{Dataset}
\label{sec:dataset_aptos}

APTOS 2019 Blindness Detection dataset~\citep{aptos2019} is used as a complementary retinal-imaging evaluation setting to evaluate class imbalance and reduced-data regimes. Specifically, a preprocessed derivation of this corpus~\citep{sovitrath2019} was employed, in which all images were resized to a uniform dimension of $224 \times 224$ pixels and underwent Gaussian filtering to reduce high-frequency noise and lighting artifacts. The dataset comprises 3,662 retinal fundus images categorized into five severity levels: No \ac{DR} (level 0), Mild \ac{DR} (level 1), Moderate \ac{DR} (level 2), Severe \ac{DR} (level 3), and Proliferative \ac{DR} (level 4). As shown in Table~\ref{tab:aptos_distribution}, the dataset exhibits significant class imbalance, with the majority of samples belonging to the No \ac{DR} category.

\begin{table}[htbp]
\centering
\caption{Distribution of fundus images across the five severity grades for the diabetic retinopathy dataset.}
\label{tab:aptos_distribution}
\begin{tabular}{lcccccc}
\toprule
Subset & No DR & Mild & Moderate & Severe & Proliferative & Total \\
\midrule
Train      & 1,444 & 296 & 799 & 155 & 235 & 2,929 \\
Validation & 180   & 37  & 100 & 19  & 30  & 366 \\
Test       & 181   & 37  & 100 & 19  & 30  & 367 \\
\midrule
Total & 1,805 & 370 & 999 & 193 & 295 & 3,662 \\
\bottomrule
\end{tabular}
\end{table}

\subsubsection{Experimental configurations for robustness evaluation}

The predictive uncertainty quantified in \acp{BNN} comprises both aleatoric and epistemic components. Aleatoric uncertainty arises from inherent noise in the data, such as image quality variations and ambiguous lesion boundaries, which cannot be reduced by collecting more data. Epistemic uncertainty, in contrast, stems from insufficient knowledge due to limited training data or model capacity, and is theoretically reducible by observing more data or increasing model capacity~\citep{gal2016uncertainty}.

Since the same dataset is utilized with only the sample size varying while other factors remain constant, the observed differences in uncertainty across configurations primarily reflect changes in epistemic uncertainty, as the aleatoric component is inherent to the data distribution and thus theoretically invariant.

To evaluate the robustness of the proposed training-time $\mathcal{L}_{\text{CUB}}$ and post-hoc \ac{DTS} under realistic deployment conditions and analyze their behavior under varying levels of epistemic uncertainty, experiments were conducted using the following data configurations:

\begin{itemize}
    \item Full dataset (100\%): the original dataset with inherent class imbalance, partitioned into training, validation, and test subsets using a stratified split of 80:10:10.

    \item Reduced train/val (50\%): only 50\% of the original training and validation samples are randomly retained, while the test set remains identical to that of the full dataset. Class proportions are preserved within the reduced subsets.

    \item Reduced train/val (25\%): only 25\% of the original training and validation samples are retained under the same protocol, representing an extreme data scarcity scenario with a fixed test distribution.

    \item Balanced (train/val): the original test set is isolated first to preserve the natural class distribution for unbiased evaluation. The training and validation sets are then balanced via data augmentation applied only to minority classes.
\end{itemize}

\subsubsection{Model architecture and training:}

For this experiment, the ResNet-34 architecture was chosen as the deterministic backbone. This selection is substantiated by extensive comparative analyses in the literature. As demonstrated in~\citep{bhimavarapuAutomaticDetectionClassification2023}, ResNet architectures consistently outperform shallower networks in \ac{DR} detection by effectively mitigating the vanishing gradient problem through residual skip connections. This depth is crucial to capture the details necessary to identify early-stage \ac{DR} lesions. Furthermore, studies on transfer learning for medical imaging~\citep{khalifaDeepTransferLearning2019, alwakidEnhancingDiabeticRetinopathy2023} highlight that ResNet variants pre-trained on ImageNet possess robust feature extraction capabilities that generalize exceptionally well to retinal fundus images~\citep{escorcia-gutierrezAnalysisPretrainedConvolutional2022}.

\paragraph{BResNet-34:}

To adapt the architecture for the \ac{DR} severity grading task, the original fully connected layer was replaced with a custom classification head. The modified head consists of a linear layer reducing the 512-dimensional feature vector to 128 dimensions, followed by a ReLU activation and dropout regularization (rate = 0.1) for overfitting prevention, and a final linear layer mapping to the output classes. The modified network was subsequently converted to a Bayesian variant, denoted \mbox{BResNet-34}, using the Bayesian-Torch library~\citep{krishnan2022bayesiantorch}.

\paragraph{Training protocol and hyperparameters:}

The BResNet was then trained using the combined objective (Eq.~\ref{eq:total_loss}), with the same \ac{SGD} optimizer, cosine annealing schedule, and \ac{MC} sampling configuration used in Experiment~1. Training required 80 epochs and a batch size of 16. A warm-up strategy was used with $\beta$ set to 0 for the first 5 epochs, and 1.2 for the remaining.
For preprocessing, retinal fundus images were normalized using standard ImageNet statistics (mean = [0.485, 0.456, 0.406], std = [0.229, 0.224, 0.225]) for each channel. Data augmentation was applied during training, including random horizontal flipping (probability = 0.5), random rotation (up to 10 degrees), and color jittering (brightness and contrast variation of 0.1). During validation and testing, only normalization was applied without augmentation.

\paragraph{Post-hoc calibration protocol:}
The same \ac{DTS} post-hoc calibration protocol used in Experiment~1 was applied to the \ac{CUB-Loss}-trained BResNet model, using the held-out validation set with frozen weights.

\subsubsection{Experimental results}

This section reports the results obtained for the Trainig-time and the Post-hoc Calibration in Experiment 2.

\paragraph{Training-Time Calibration:}

\begin{table}[htbp]
\centering
\caption{Diabetic retinopathy detection results.}
\label{tab:dr_5class}
\begin{tabular}{lcccc}
\toprule
\multirow{2}{*}{Data Configuration} & \multicolumn{2}{c}{BResNet} & \multicolumn{2}{c}{+ $\mathcal{L}_{\text{CUB}}$} \\
\cmidrule(lr){2-3} \cmidrule(lr){4-5}
 & Acc. (\%) & AvU & Acc. (\%) & AvU \\
\midrule
Reduced train/val (25\%) & 70 & 0.68 & 76 & 0.76 \\
Reduced train/val (50\%) & 74 & 0.70 & 79 & 0.73 \\
Full dataset             & 80 & 0.68 & 81 & 0.76 \\
Balanced (train/val)     & 82 & 0.61 & \textbf{82} & \textbf{0.79} \\
\bottomrule
\end{tabular}
\vspace{0.5cm}
\caption{Mean uncertainty for correct and incorrect predictions under different data configurations.}
\label{tab:dr_uncertainty_detail}
\begin{tabular}{lcccc}
\toprule
\multirow{2}{*}{Data Configuration} & \multicolumn{2}{c}{BResNet} & \multicolumn{2}{c}{+ $\mathcal{L}_{\text{CUB}}$} \\
\cmidrule(lr){2-3} \cmidrule(lr){4-5}
 & $U_{\text{correct}}$ & $U_{\text{incorrect}}$ & $U_{\text{correct}}$ & $U_{\text{incorrect}}$ \\
\midrule
Reduced Train/Val (25\%) & 0.56 & 1.20 & 0.26 & 0.82 \\
Reduced Train/Val (50\%) & 0.48 & 1.14 & 0.26 & 0.81 \\
Full Dataset (100\%)     & 0.34 & 0.85 & 0.19 & 0.65 \\
Balanced (Train/Val)     & 0.46 & 1.01 & 0.16 & 0.56 \\
\bottomrule
\end{tabular}
\end{table}

Table~\ref{tab:dr_5class} and Table~\ref{tab:dr_uncertainty_detail} summarize the detection performance and uncertainty estimates under different training/validation data configurations with a fixed test distribution. 

As shown in Table~\ref{tab:dr_5class}, the proposed $\mathcal{L}_{\text{CUB}}$ improves both detection accuracy and uncertainty calibration across all settings. Under the balanced (train/val) configuration where both models achieve identical accuracy (82\%), $\mathcal{L}_{\text{CUB}}$ yields a substantially higher \ac{AvU} (0.79 vs 0.61), demonstrating that the calibration improvement is independent of accuracy gains. 

Unlike Experiment 1, where the abundant training data and balanced class distribution effectively reduce epistemic uncertainty affection, making $\Delta U$ a fair metric for comparing prediction performance, Experiment 2 does not adopt this metric. Due to the limited availability and class imbalance in this dataset, the baseline BResNet exhibits sensitivity to epistemic uncertainty, resulting in uncalibrated uncertainty outputs. Under such conditions, the $\Delta U$ comparison becomes unreliable.

Instead, the analysis focuses on demonstrating the robustness of $\mathcal{L}_{\text{CUB}}$ under elevated epistemic uncertainty. Under severe data scarcity (25\% training data), the baseline BResNet exhibits substantially increased uncertainty for both correct and incorrect predictions, indicating high sensitivity to epistemic uncertainty arising from limited training data. In contrast, $\mathcal{L}_{\text{CUB}}$ not only maintains higher accuracy (76\% vs 70\%) but also preserves stable uncertainty estimates, with $U_{\text{correct}}$ remaining consistently low across all configurations. According to the definition of epistemic uncertainty, this ability to maintain calibration under data scarcity demonstrates that $\mathcal{L}_{\text{CUB}}$ effectively reduces the impact of epistemic uncertainty, indicating stronger model capability. 

The balanced configuration further validates this finding. Data balancing through augmentation effectively increases the amount of training data for minority classes, thereby reducing epistemic uncertainty. Under this configuration, $\mathcal{L}_{\text{CUB}}$ achieves the lowest uncertainty for both correct predictions ($U_{\text{correct}}$ = 0.16) and incorrect predictions ($U_{\text{incorrect}}$ = 0.56) across all settings, demonstrating that the proposed method can fully leverage the reduced epistemic uncertainty from data augmentation to achieve optimal calibration performance.

These results demonstrate that the proposed $\mathcal{L}_{\text{CUB}}$ effectively enhances the calibration capability of \acp{BNN}, providing more robust uncertainty estimates in data-scarce scenarios by mitigating the impact of epistemic uncertainty arising from limited training data. 

\paragraph{Post-hoc Calibration:}

Table~\ref{tab:dr_posthoc_boundary} shows the metrics used to evaluate the post-hoc calibration for diabetic retinopathy. The post-hoc pattern for Experiment 2 is consistent with that obtained in Experiment~1. \acs{DTS} reduces \acs{BCCE} across all data configurations while leaving the reported accuracy unchanged, and \acs{AvU} is only marginally affected, consistent with the preservation property described in Section~\ref{sec:dts_algorithm}: post-hoc temperature scaling reshapes the confidence--uncertainty profile without altering the learned decision structure. As a dual-temperature calibration strategy, \acs{DTS} may cause these complementary diagnostics to shift in either direction, a behavior consistent with Experiment~1 and reflecting differences in calibration mechanisms and optimization directions. The $U_{\text{correct}}$ and $U_{\text{incorrect}}$ columns further show that \acs{DTS} refines the uncertainty profile on top of \ac{CUB-Loss} training toward a more stable separation pattern rather than enforcing a uniform increase or decrease in uncertainty. After \acs{DTS}, the uncertainty estimates exhibit a stable decreasing trend across data configurations, indicating lower uncertainty as the available training data increase. This pattern indicates that the post-hoc calibration reinforces, rather than disrupts, the uncertainty structure learned during training, producing more reliable uncertainty estimates that better reflect the confidence--uncertainty relationship defined by the \acs{CUBC} boundary target.

\begin{table*}[htbp]
\centering
\caption{Post-hoc calibration for diabetic retinopathy.}
\label{tab:dr_posthoc_boundary}
\resizebox{\textwidth}{!}{
\begin{tabular}{llccccccc}
\toprule
Data Configuration & Method & Acc. (\%) $\uparrow$ & ECE (\%) & UCE (\%) & BCCE (\%) $\downarrow$ & AvU $\uparrow$ & $U_{\text{correct}}$ & $U_{\text{incorrect}}$ \\
\midrule
\multirow{2}{*}{Reduced train/val (25\%)} & + $\mathcal{L}_{\text{CUB}}$ & 76 & 9.52 & 4.84 & 8.09 & 0.76 & 0.26 & 0.82 \\
 & + $\mathcal{L}_{\text{CUB}}$ + DTS & 76 & 7.85 & 5.37 & 7.76 & 0.75 & 0.29 & 0.89 \\
\midrule
\multirow{2}{*}{Reduced train/val (50\%)} & + $\mathcal{L}_{\text{CUB}}$ & 79 & 7.84 & 5.56 & 8.48 & 0.73 & 0.26 & 0.81 \\
 & + $\mathcal{L}_{\text{CUB}}$ + DTS & 79 & 6.28 & 8.24 & 7.50 & 0.73 & 0.27 & 0.88 \\
\midrule
\multirow{2}{*}{Full dataset} & + $\mathcal{L}_{\text{CUB}}$ & 81 & 8.62 & 5.47 & 7.37 & 0.76 & 0.19 & 0.65 \\
 & + $\mathcal{L}_{\text{CUB}}$ + DTS & 81 & 8.55 & 4.92 & 7.22 & 0.76 & 0.19 & 0.66 \\
\midrule
\multirow{2}{*}{Balanced (train/val)} & + $\mathcal{L}_{\text{CUB}}$ & 82 & 9.31 & 5.82 & 6.91 & 0.79 & 0.16 & 0.56 \\
 & + $\mathcal{L}_{\text{CUB}}$ + DTS & 82 & 9.49 & 6.37 & 6.69 & 0.80 & 0.16 & 0.57 \\
\bottomrule
\end{tabular}
}
\end{table*}

\subsection{Experiment 3: identification of skin lesions}

The visual differentiation between malignant melanomas and benign pigmented lesions is challenging even for experienced dermatologists due to high inter-class similarity and visual complexity. While deep learning has shown promise in automating this process, the severe class imbalance in real-world dermatological data poses a significant risk of model bias. Therefore,  calibration mechanisms that can accurately quantify the uncertainty for underrepresented classes are essential for a reliable clinical decision~\citep{fernandoDynamicallyWeightedBalanced}.

\subsubsection{Dataset}
\label{sec:dataset_ham10000}

The proposed method was evaluated using the \ac{HAM10000} dataset ~\citep{tschandl2018ham10000}, a large-scale collection of multi-source dermatoscopic images of common pigmented skin lesions. The original dataset contains multiple images of the same lesion captured at different magnifications or angles. To ensure a rigorous evaluation and prevent data leakage between training and testing sets, a cleaning procedure was performed to remove duplicate entries derived from the same lesion identifier. This resulted in a refined corpus of 7,470 unique dermatoscopic images.

The dataset contains seven different categories: \ac{NV}, \ac{MEL}, \ac{BKL}, \ac{BCC}, \ac{AKIEC}, \ac{VASC}, and \ac{DF}. As illustrated in the distribution statistics, the data exhibit extreme class imbalance; the majority class (\ac{NV}) accounts for approximately 72.3\% of the total dataset (5,403 images), whereas rare classes like \ac{DF} contain as few as 73 images.

 A stratified splitting strategy was followed to partition the dataset into training, validation, and test subsets with a ratio of 80:10:10. This ensures that the distinct relative proportions of each category are preserved across all folds. The specific distribution of the images per class and subset is detailed in Table ~\ref{tab:ham10000_distribution}.

\begin{table}[htbp]
\centering
\caption{Distribution of skin lesions across the seven categories in the HAM10000 dataset.}
\label{tab:ham10000_distribution}
\begin{tabular}{lcccc}
\toprule
Dataset Category & Train & Validation & Test & Total \\
\midrule
\acs{NV} & 4,322 & 540 & 541 & 5,403 \\
\acs{BKL} & 581 & 72 & 74 & 727 \\
\acs{MEL} & 491 & 61 & 62 & 614 \\
\acs{BCC} & 261 & 32 & 34 & 327 \\
\acs{AKIEC} & 182 & 22 & 24 & 228 \\
\acs{VASC} & 78 & 9 & 11 & 98 \\
\acs{DF} & 58 & 7 & 8 & 73 \\
\midrule
Total & 5,973 & 743 & 754 & 7,470 \\
\bottomrule
\end{tabular}
\end{table}

\subsubsection{Model architecture and training}

For this experiment, the ResNet-50 architecture was chosen as the deterministic backbone. According to the systematic review by~\citep{debeleeSkinLesionClassification2023}, ResNet-50 offers a favorable trade-off between computational complexity and predictive performance for the identification of skin lesions, making it a particularly suitable choice for this task compared with alternative backbone architectures.

\paragraph{BResNet-50:}

The architecture for the seven-class skin lesion classification task was adapted, replacing the original fully connected layer with a custom classification head consisting of a linear layer reducing the 2048-dimensional feature vector to 512 dimensions, followed by a ReLU activation and dropout regularization (rate = 0.3), and a final linear layer mapping to the seven output classes. The modified network was subsequently converted to a Bayesian variant, denoted \mbox{BResNet-50}, using the Bayesian-Torch library~\citep{krishnan2022bayesiantorch}.

\paragraph{Training protocol and hyperparameters:}

The same \ac{MC} sampling configuration followed in Experiment~1 was adopted. Given the severe class imbalance in the \ac{HAM10000} dataset, class-weighted cross-entropy loss was employed in both training stages to ensure proportional minority-class contributions, with class weights computed using the balanced weighting scheme.

The BResNet-50 was then trained using the combined objective (Eq.~\ref{eq:total_loss}), with the same \ac{SGD} momentum and weight decay settings as in Experiment~1, a learning rate of $1\times10^{-3}$, and a batch size of 64. A warm-up strategy was used with $\beta$ set to 0 for the first 30 epochs, and fixed to 0.6 for the remaining; the extended warm-up period was used to stabilize Bayesian training before introducing uncertainty calibration. Early stopping based on \ac{BACC} with a patience of 15 consecutive epochs was applied after the warm-up phase.

\paragraph{Post-hoc calibration protocol:}
The same \ac{DTS} post-hoc calibration protocol as in Experiment~1 was applied to the \ac{CUB-Loss}-trained BResNet-50, using the held-out validation set with frozen weights.

\subsubsection{Experimental results}

The results in this section are presented first for the trainig-
time calibration and later for the post-hoc Calibration.

\paragraph{Training-Time Calibration:}

Table~\ref{tab:skin_results} presents the training-time calibration results on the \ac{HAM10000} dataset. The severe class imbalance makes this setting particularly challenging for the Bayesian uncertainty estimation. The BResNet-50 baseline achieves an accuracy of 81\%, a \ac{BACC} of 64\%, and an \ac{AvU} of 0.79. The addition of $\mathcal{L}_{\text{CUB}}$ improves both predictive performance and uncertainty alignment, also increasing the accuracy up to 85\%, \ac{BACC} to 66\%, and \ac{AvU} to 0.84. These results indicate that $\mathcal{L}_{\text{CUB}}$ can strengthen training-time calibration for Bayesian models under severe class imbalance, producing uncertainty estimates that better reflect prediction reliability in challenging settings as the one present in Experiment 2.

\begin{table}[htbp]
\centering
\caption{Results on the identification of skin lesions using the HAM10000 dataset.}
\label{tab:skin_results}
\begin{tabular}{lccc}
\toprule
Method & Acc. (\%) & BACC (\%) & AvU \\
\midrule
BResNet-50 (Baseline) & 81 & 64 & 0.79 \\
BResNet-50 + $\mathcal{L}_{\text{CUB}}$ (Ours) & 85 & 66 & \textbf{0.84} \\
\bottomrule
\end{tabular}
\end{table}

\paragraph{Post-hoc Calibration:}

\begin{table*}[htbp]
\centering
\caption{Post-hoc calibration on HAM10000.}
\label{tab:ham_posthoc_boundary}
\begin{tabular}{lcccccc}
\toprule
Method & Acc. (\%) $\uparrow$ & BACC (\%) $\uparrow$ & ECE (\%) & UCE (\%) & BCCE (\%) $\downarrow$ & AvU $\uparrow$ \\
\midrule
BResNet-50 + $\mathcal{L}_{\text{CUB}}$ & 85 & 66 & 9.25 & 6.11 & 4.23 & 0.84 \\
BResNet-50 + $\mathcal{L}_{\text{CUB}}$ + DTS & 85 & 66 & 7.54 & 9.59 & 3.92 & 0.84 \\
\bottomrule
\end{tabular}
\end{table*}

Table~\ref{tab:ham_posthoc_boundary} further shows that \acs{DTS} remains applicable to the \acs{CUB-Loss}-trained model under severe class imbalance. With the model weights fixed, \acs{DTS} preserves accuracy, \acs{BACC}, and \acs{AvU}, while further reducing \acs{BCCE} from 4.23\% to 3.92\%. Despite the increase in \acs{UCE}, the lower \acs{BCCE} indicates that \acs{DTS} can refine the confidence--uncertainty boundary relationship after training-time calibration, supporting its role as a post-hoc boundary-alignment adjustment for imbalanced medical image classification.

\section{Discussion}\label{sec:discussion}

Uncertainty has been explored in medical image analysis to identify cases that may benefit from secondary expert review, selective referral, or additional diagnostic attention~\citep{leibig2017leveraging,combalia2020uncertainty}. In such expert support settings, a scalar uncertainty score becomes more informative when its expected range is anchored to the confidence level of the prediction, since this anchoring allows relative differences in uncertainty across samples to be interpreted consistently rather than read in isolation. The proposed \ac{CUBC} provides such an anchor by defining the expected confidence--uncertainty relationship, while \ac{BCCE} quantifies deviations from this boundary and the two calibration components, \ac{CUB-Loss} and \ac{DTS}, align predictive uncertainty with it during and after training.

Building on this anchor-based interpretation, the three experiments presented provide complementary evidence that uncertainty anchored by \ac{CUBC} is useful across different medical imaging applications and settings with different data and deployment characteristics. Experiment 1 shows that the calibrated uncertainty score can simultaneously support the uncertainty-aware screening task and provide a useful check for near \ac{OOD} cases, indicating that the same representation aligned with the boundary can serve both uncertainty estimation and awareness of distribution shift. Experiment 2 evaluates robustness under data scarcity and class imbalance, showing that $\mathcal{L}_{\text{CUB}}$ provides implicit regularization by maintaining higher accuracy and a more stable \ac{AvU} under reduced training data. In the same setting, \ac{DTS} further aligns the uncertainty estimates with the trend induced by the available training data, making the calibrated uncertainty vary more consistently as the data regime changes. Experiment 3 further shows that the boundary alignment principle transfers to a more imbalanced Bayesian setting, where better task-level discrimination and better uncertainty calibration can be achieved together. Collectively, these results indicate that the main benefit of \ac{CUBC} calibration lies in making the confidence--uncertainty relationship more stable across changes in task setting, data availability, and class balance.

The proposed framework opens several directions for future investigation. First, collaborating with clinical partners to evaluate the framework on datasets with rich patient-level annotations and clearly defined diagnostic semantics would enable a more grounded analysis of how confidence--uncertainty boundary alignment relates to clinically meaningful uncertainty. Second, integrating clinician feedback into the calibration process, for example, by using expert disagreement or referral decisions as signals for boundary refinement, would bring the framework closer to real-world clinical decision support. Third, extending \ac{CUB-Loss} and \ac{DTS} to more model families and broader uncertainty methods would clarify how widely the boundary alignment principle transfers across modeling paradigms.

\section{Conclusions}\label{sec:conclusions}

This paper presents a confidence--uncertainty boundary calibration framework for Bayesian medical image classification, comprising two complementary components: the \ac{CUB-Loss} ($\mathcal{L}_{\text{CUB}}$) for training-time calibration, and \ac{DTS} for post-hoc refinement. Across retrospective evaluations covering pneumonia screening under both \ac{ID} and near \ac{OOD} settings, \ac{DR} detection under data scarcity, and skin lesion identification under class imbalance, \ac{CUB-Loss} promotes correctness-aware uncertainty behavior during learning, while \acs{DTS} further improves post-hoc \ac{CUBC} alignment and preserves Accuracy and \acs{AvU} in the evaluated settings. By linking training-time and post-hoc boundary calibration, the results suggest a practical route for extending uncertainty-aware Bayesian models toward more reliable real-world medical image analysis.

\section*{Data and code availability}

All datasets used in this study are publicly available:

\begin{itemize}
    \item Pneumonia screening: compiled from MIMIC-CXR (\url{https://physionet.org/content/mimic-cxr/2.1.0/}), CheXpert (\url{https://stanfordaimi.azurewebsites.net/datasets/8cbd9ed4-2eb9-4565-affc-111cf4f7ebe2}), COVIDx CXR-2 (\url{https://www.kaggle.com/datasets/andyczhao/covidx-cxr2}), and NIH Chest X-rays (\url{https://www.kaggle.com/datasets/nih-chest-xrays/data}). Dataset construction details are described in the paper.
    
    \item Diabetic retinopathy detection: preprocessed dataset available at \url{https://www.kaggle.com/datasets/sovitrath/diabetic-retinopathy-224x224-gaussian-filtered}, derived from the APTOS 2019 Blindness Detection challenge.
    
    \item Identification of skin lesions: available at \url{https://www.kaggle.com/datasets/kmader/skin-cancer-mnist-ham10000}.
\end{itemize}

The code is publicly available at \url{https://github.com/BYO-UPM/CUB-Loss}.

\section*{CRediT authorship contribution statement}

Hua Xu: Writing – original draft, Software, Methodology, Investigation, Formal analysis, Data curation, Conceptualization.

Julián D. Arias-Londoño: Writing – review \& editing, Supervision, Methodology, Formal analysis, Conceptualization.

Juan I. Godino-Llorente: Writing – review \& editing, Supervision, Resources, Project administration, Methodology, Funding acquisition, Formal analysis, Conceptualization.

\section*{Declaration of generative AI and AI-assisted technologies in the writing process}

During the preparation of this manuscript, the authors used ChatGPT (OpenAI) to enhance the quality of the English language and improve the overall clarity and style of the text. All outputs generated by this tool were carefully reviewed and, where necessary, revised by the authors, who assume full responsibility for the final content of the published work.

\section*{Declaration of competing interest}
The authors declare that they have no known competing
financial interests or personal relationships that could appear
to influence the work reported in this document.

\section*{Acknowledgments}
\begin{sloppypar}
This work was supported by the Ministry of Economy and Competitiveness of Spain under grants DPI2017-83405-R1 and PID2021-128469OB-I00, and by Comunidad de Madrid, Spain. Universidad Politécnica de Madrid supports Julián D. Arias-Londoño through a María Zambrano UPM2021-035 grant funded by European Union-NextGenerationEU. Hua Xu acknowledges the support provided by the China Scholarship Council. 
The authors also thank the Madrid ELLIS unit (European Laboratory for Learning \& Intelligent Systems) for its indirect support.
Finally, the authors acknowledge Universidad Politécnica de Madrid for providing computing resources on the Magerit Supercomputer.
\end{sloppypar}

\bibliographystyle{unsrt}  
\bibliography{references}

\end{document}